\documentclass{article}

\usepackage{iclr2027_conference,times}
\iclrfinalcopy

\usepackage{amsmath,amsfonts,bm}

\def\eqref#1{equation~\ref{#1}}

\def\1{\bm{1}}

\DeclareMathAlphabet{\mathsfit}{\encodingdefault}{\sfdefault}{m}{sl}
\SetMathAlphabet{\mathsfit}{bold}{\encodingdefault}{\sfdefault}{bx}{n}

\usepackage{float}
\usepackage{hyperref}
\hypersetup{hidelinks}
\usepackage{url}
\usepackage{graphicx}
\usepackage{booktabs}
\usepackage{subcaption}
\usepackage{amsmath}
\usepackage{hyperref}
\usepackage{fontawesome5}
\usepackage{placeins}

\title{\raggedright SAGE: Subgoal-Conditioned Action Generation for Latent World Model Planning}

\author{
  {\bf Letian Cheng}\thanks{Equal contribution} \qquad
  {\bf Qi Zhang}\footnotemark[1] \qquad
  {\bf Yisen Wang}\thanks{Corresponding author: Yisen Wang (yisen.wang@pku.edu.cn)} \\
  \ \ Peking University\\
}

\begin{document}

\maketitle
\fancyhead[L]{}

\begin{center}
  \href{https://github.com/PKU-ML/SAGE}
       {\faGithub\ \textbf{Code}}
  \qquad
  \href{https://huggingface.co/CLTRAY/SAGE}
       {\faCube\ \textbf{Model}}
\end{center}

\begin{abstract}

Latent world models have emerged as a powerful planning paradigm by learning action-conditioned predictive dynamics and using them as internal simulators to imagine and evaluate candidate action sequences. However, as the planning horizon grows, performance becomes increasingly constrained by proposal quality: a fixed candidate budget must search an exponentially larger action space, making it difficult to expose the world model to high-quality candidate futures for evaluation. In this paper, we introduce SAGE, a prior-conditioned planner that replaces random proposal initialization with structured guidance. At each planning stage, a goal-conditioned generator predicts the next intermediate latent subgoal for a specified duration, which is then used to condition the generation of candidate action sequences. To capture semantic information across temporal scales, we use subgoals of varying durations as priors, balancing fine-grained local control with higher-level long-horizon progress. Then the frozen world model evaluates these proposals against the same subgoal and guides their refinement before execution. Experiments on PushT and OGBench Cube show that coupling latent subgoal decomposition with prior-conditioned action generation substantially improves long-horizon planning while preserving strong short-horizon performance. To be specific, when the target offset is $150$, it raises PushT success from $4.7\%$ to $64.7\%$ and OGBench Cube success from $20.7\%$ to $67.3\%$. We further extend latent world-model planning to LIBERO, where SAGE improves full-episode success from $0\%$ with the vanilla LeWM planner to $48.7\%$ on Scene2 and $58\%$ on Caddy.

\end{abstract}

\section{Introduction}

Latent world models offer a compelling framework for planning by learning how latent representations evolve under candidate actions and using these learned dynamics to simulate possible futures. Instead of reconstructing future observations at the pixel level, they predict directly in a compact representation space and evaluate imagined outcomes against a target, making planning substantially more efficient \citep{dinowm2025,lewm,vjepa2}. During the planning process, a typical planner repeatedly proposes action sequences, rolls them forward through the latent dynamics, and selects the candidates whose predicted futures best match the desired goal.

The planning process naturally separates prediction from proposal: the world model determines how candidate actions are evaluated, whereas the proposal mechanism determines which parts of the action space are exposed to that evaluation \citep{dinowm2025}. As the planning horizon grows, the proposal mechanism becomes increasingly critical \citep{prism2026}. Under a fixed candidate budget, initializing action search from a generic random distribution provides increasingly poor coverage of the rapidly expanding action space, making goal-directed action sequences unlikely to be sampled. A more effective planner therefore requires a structured proposal distribution that concentrates search on plausible behaviors. Short action sequences also need a suitable intermediate target: useful preparatory motions may make little immediate progress toward the final goal.

Consequently, in this paper, we introduce SAGE, a prior-conditioned planning framework in which actions are sampled conditioned on predicted latent subgoals rather than initialized from a generic random distribution. To be specific, we first train a variable-length latent subgoal generator to decompose long-horizon tasks into different temporal scales. Given the current observation history, the distant goal, and a requested future horizon, the generator predicts one intermediate latent target at that temporal offset. Shorter subgoals preserve fine-grained information needed for local corrections, whereas longer subgoals capture higher-level progress toward the final goal. A single generator supports multiple subgoal lengths, allowing the same planning framework to reason over both local and coarse temporal abstractions. 

We then use these generated subgoals as structured priors for action search. Rather than initializing the planner from a generic random distribution, the planner generates action sequences conditioned on these predicted subgoals over the corresponding horizon. The frozen world model evaluates the predicted outcomes against the same subgoal and refines the proposals before execution. In this way, variable-length subgoals not only decompose the distant goal, but also directly shape the candidate distribution explored by the planner.

Experiments on PushT and OGBench Cube validate our method under a controlled planning setup. Using the same frozen LeWM, candidate budget, and CEM refinement procedure as Base LeWM, our planner improves success across planning horizons, with particularly large gains on distant goals. Component studies further show that both generating local targets and using them as proposal priors yield substantial gains, with additional gains from LeWM-guided selection and refinement. We further evaluate SAGE on the more realistic and complex LIBERO manipulation benchmark. Under matched execution budgets, SAGE raises full-episode success from 0\% with the base LeWM planner to 48.7\% on Scene2 and 58.0\% on Caddy.

The contributions can be summarized as:
\begin{itemize}
\item We introduce a multi-horizon latent subgoal generator that predicts intermediate latent targets at flexible temporal offsets, enabling the planner to capture both local dynamics and longer-term task progress.
\item We develop a subgoal-conditioned action proposal mechanism and consistently use the predicted subgoal to guide action generation, proposal evaluation, and refinement with a frozen latent world model.
\item We demonstrate improved planning on PushT and OGBench Cube, and extend the improvement to full-episode LIBERO manipulation through scalable oracle-driven data collection, with extensive component ablations.
\end{itemize}

\section{Related Work}

\subsection{Latent World Model Planning}
Latent world models support control by predicting how compact state representations evolve under actions and using imagined trajectories to guide behavior.Model-based reinforcement learning methods exploit learned latent
dynamics in different ways: Dreamer learns policies and value functions from imagined trajectories \citep{dreamerv3}, whereas TD-MPC2 combines latent-space model-predictive control with learned value estimates \citep{tdmpc2}. A complementary line considers reward-free, goal-conditioned planning, where candidate actions are optimized by comparing predicted latent states against the latent representation of the goal observation. PLDM learns JEPA-based latent dynamics from offline, reward-free trajectories and studies their generalization for zero-shot goal reaching \citep{pldm2025}. DINO-WM instead predicts future DINO patch features and plans by treating target features as the optimization objective \citep{dinowm2025}, while LeWM learns the visual representation and action-conditioned dynamics jointly from pixels before planning with CEM \citep{lewm}. V-JEPA 2 scales joint-embedding prediction through large-scale video pretraining and adapts the resulting model to action-conditioned prediction and robotic planning \citep{ijepa,vjepa2}. We keep the encoder, latent dynamics, and latent distance function fixed, and improve the local targets and action proposals used for planning.

\subsection{Long-Horizon Planning}
Long-horizon planning is a central challenge for sequential decision-making, as agents must maintain goal-directed progress while coordinating decisions across multiple temporal scales. A broad range of approaches has therefore explored intermediate decisions and structured action distributions. The options framework formalizes temporally extended closed-loop behaviors \citep{options1999}; HIRO learns a high-level subgoal policy together with a low-level controller \citep{hiro}; and Play-LMP learns latent plans and goal-conditioned control from teleoperated play \citep{playlmp}. Guider similarly combines an offline latent subgoal prior with a reachability-aware controller \citep{guider2023}. More recent world-model approaches introduce temporal abstraction directly into predictive modeling: HWM learns dynamics at multiple temporal scales and plans hierarchically across them \citep{hwm2026}, whereas VLWM predicts latent transitions over action segments of different lengths \citep{vlwm2026}.

Two concurrent works come closest to the setting studied here. PRISM introduces learned proposal guidance for frozen JEPA planning by predicting a Gaussian action prior conditioned on the current state and integrating it into MPPI \citep{mppi,prism2026}. Our method introduces stronger task structure by first predicting a reachable latent subgoal and then conditioning an action prior on the local progress specified by that subgoal. FF-JEPA also predicts latent subgoals for successive local planning \citep{ffjepa2026}, but uses them only as optimization targets rather than as conditioning signals for learned action proposals. In our method, each subgoal jointly defines the target evaluated by latent world models and the action distribution explored by the planner. Moreover, a single generator in our framework supports multiple requested durations, enabling the same framework to coordinate planning across temporal scales.

\section{Subgoal-Conditioned Action Generation Planner for World Model Planning}
When the target lies far beyond the planner’s immediate prediction horizon, randomly generating unconstrained action candidates and ranking them according to the final goal provides quite limited guidance. We therefore decompose long-horizon planning into two coupled problems: selecting a useful and reachable subgoal, and generating an action sequence conditioned on it. Based on this idea, we introduce \textsc{SAGE} (subgoal-conditioned action generation planner), which learns both components while keeping the world model fixed. Specifically, Section \ref{sec:preliminaries} introduces the necessary preliminaries, Sections \ref{sec:subgoal-generator} and \ref{sec:goal-conditioned-planner} present the two components, and Section \ref{sec:training-components} describes the details for training and planning with these components.

\subsection{Preliminaries}
\label{sec:preliminaries}

Latent world models such as LeWM \citep{lewm} and DINO-WM \citep{dinowm2025} consist of an image encoder $E$ and an action-conditioned latent predictor $f$. Let $o_t$ denote the observation at time step $t$, with latent representation $z_t=E(o_t)$. We define the recent history $h_t$ as a sequence of past latent states and actions, i.e., $h_t=\big((z_{t-k},a_{t-k}),\ldots,(z_{t-1},a_{t-1}),z_t\big)$. During training, the predictor estimates the latent representation of the next observation conditioned on this history and the current action:
$$
\hat{z}_{t+1}=f(h_t,a_t),
$$
where $z_{t+1}=E(o_{t+1})$ provides the prediction target. In this way, $f$ learns action-conditioned dynamics directly in the latent space.

During planning, the agent additionally receives a goal image $g$, encoded as $z_g=E(g)$. We refer to an action sequence $a_{1:\tau}=(a_1,\ldots,a_\tau)$ as a candidate, which is executed open-loop before replanning. To evaluate a candidate, the world model autoregressively rolls out its action blocks to predict the terminal latent, denoted by $F(h_t,a_{1:\tau})$ . A sampling-based planner generates candidates, compares their predicted terminal latents $F(h_t,a_{1:\tau})$ with the target latent $z_g$, and selects an option to execute. After execution, the agent receives new observations and replans. SAGE keeps both $E$ and $f$ frozen and learns two additional components: a subgoal generator and an action prior.

\subsection{Goal-Conditioned Subgoal Generation}
\label{sec:subgoal-generator}

The standard planning procedure above relies on the final goal to evaluate candidate actions. However, given only a final goal, it may be unclear what the agent should achieve in the next few steps. For example, in a pushing task, the agent may first need to move to a suitable side of the object before pushing it toward the target. This intermediate state may still be far from the final goal, but it is an important step toward reaching it. Consequently, in our framework, we first generate a subgoal latent that provides a local target for evaluating action candidates at the current planning step.

Specifically, we denote the subgoal generator as $G$. To determine where this subgoal should lie, we condition the generator on two temporal quantities. The candidate duration $\tau$ denotes the number of environment steps covered by the current action candidate sequence, while the remaining goal horizon $\Delta$ denotes the number of steps to the final goal. The generator predicts the intermediate target as:
\begin{equation}
z_s^{t,\Delta,\tau}=G(h_t,z_g,\Delta,\tau).
\label{eq}
\end{equation}
The output $z_s^{t,\Delta,\tau}$ has the same token structure as an encoded goal image and is supervised by the observation $z_{t+\tau}$ in the recorded trajectory. We implement $G$ as a Transformer decoder that conditions on the history and goal tokens, together with embeddings of $\Delta$, $\tau$, and the optional state input $x$.  Additional architectural details are provided in Appendices~\ref{app:implementation} and~\ref{app:libero-training}.

\begin{figure*}[t]
  \centering
  \includegraphics[width=0.9\textwidth]{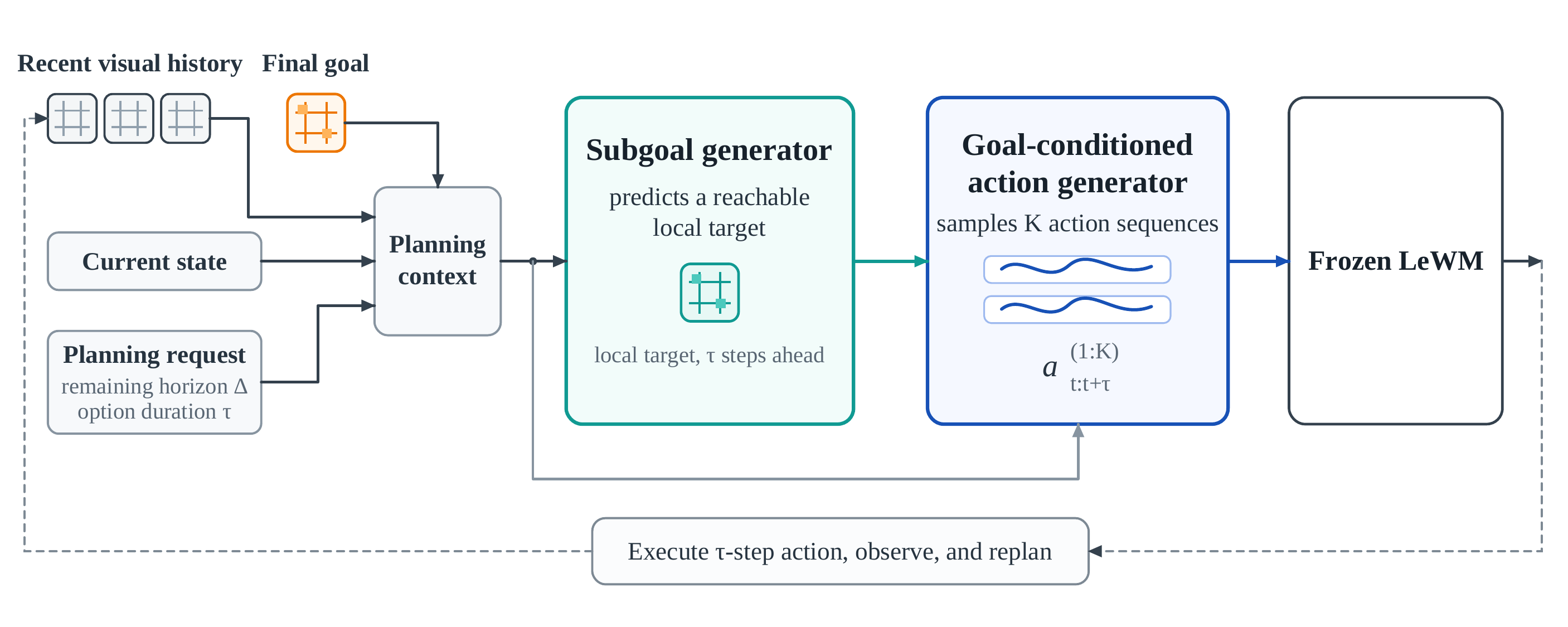}
  \caption{The planning stage in \textsc{SAGE}. The subgoal generator converts
  the current history, a far-goal latent, and a requested duration $\tau$ into
  a local target. Then the action generator proposes $\tau$-step action options
  conditioned on that target. Frozen LeWM predicts and ranks their latent
  futures, while CEM refines the best proposal region before execution.
  Each execution in the loop is a complete $\tau$-step option.}
  \label{fig:method-overview}
\end{figure*}

\subsection{Subgoals for Action Proposals and Evaluation}
\label{sec:goal-conditioned-planner}

A subgoal specifies the local state that the planner would like to reach, but it does not directly tell the agent which actions can achieve it. In principle, the planner could randomly sample action sequences and use the world model to test whether they reach the subgoal. In practice, however, most randomly sampled sequences are unlikely to produce meaningful behaviors, making the search inefficient. We therefore learn an action prior from recorded trajectories to generate plausible action candidates conditioned on the current history and the predicted subgoal. The frozen world model is then used to evaluate which of these candidates is most likely to reach the subgoal.

\paragraph{Proposing actions.}
We denote the action generator by $q$. Given the current history $h_t$, final-goal state $z_g$, temporal parameters $(\Delta,\tau)$, and predicted subgoal $z_s^{t,\Delta,\tau}$, it models a distribution over $\tau$-step action sequences $a_{1:\tau}$. Here, $z_g$ provides the global task context, while $z_s^{t,\Delta,\tau}$ specifies the local state that the candidate sequence should reach. We parameterize this distribution with a Transformer decoder and use a Gaussian mixture to capture multiple action sequences that may lead to the same subgoal:
\begin{equation}
q\left(a_{1:\tau}\mid h_t,z_g,\Delta,\tau,z_s^{t,\Delta,\tau}\right)
=
\sum_{m=1}^{M}\pi_m
\mathcal{N}\left(\mu_m,\Sigma_m\right).
\label{eq:gmm-option}
\end{equation}
Each mixture component represents one possible mode of reaching the subgoal, with probability $\pi_m$, learned mean value $\mu_m$, and diagonal covariance $\Sigma_m$. To generate a candidate, we first sample a component according to $\pi_m$ and then sample the entire action sequence from the corresponding Gaussian. In this way, all actions in a candidate sequence are generated from the same mode, preserving temporal consistency across the sequence. Additional architectural details are provided in Appendix~\ref{app:implementation}.

\paragraph{Evaluating candidate outcomes.}
The action prior proposes plausible candidates, but it does not guarantee that a sampled sequence will actually reach the predicted subgoal. We therefore evaluate each candidate using the frozen world model. Given a candidate $a_{1:\tau}$, we recursively roll out the predictor $f$ from the current history $h_t$ and obtain its predicted terminal latent $F(h_t,a_{1:\tau})$. We then measure how close this predicted outcome is to the subgoal:
\begin{equation}
C\left(a_{1:\tau};z_s^{t,\Delta,\tau}\right)
=
d\left(F(h_t,a_{1:\tau}),z_s^{t,\Delta,\tau}\right),
\label{eq:planning-cost}
\end{equation}
where $d$ denotes the distance in the backbone latent space, and a lower cost indicates that the candidate is predicted to end closer to the desired subgoal. The action prior therefore narrows the search to plausible behaviors, while the world model further distinguishes among them according to their predicted progress toward the subgoal. These candidates are then used to initialize the online search and are selected or refined using the same cost.

\subsection{Training and Planning with SAGE}
\label{sec:training-components}

The previous two sections introduced the two learned components of SAGE: a subgoal generator and a subgoal-conditioned action prior. We now describe how these two components are trained from recorded trajectories and how they are combined with the frozen world model during online planning.
\paragraph{Subgoal generator training.}
Each training sample contains a recent history $h_t$, a final goal located $\Delta$ environment steps ahead, and a recorded action sequence of duration $\tau$, where $\tau\leq\Delta$. The latent representation of the observation reached after these $\tau$ steps, denoted by $z_{t+\tau}$, serves as the supervision target for the subgoal generator. Given $(h_t,z_g,\Delta,\tau)$, the generator predicts $z_s^{t,\Delta,\tau}$ and is optimized with
\begin{equation}
\mathcal{L}_{G}
=
\operatorname{SmoothL1}
\left(
z_s^{t,\Delta,\tau},
z_{t+\tau}
\right)
+
\lambda_G
\left[
1-
\cos
\left(
z_s^{t,\Delta,\tau},
z_{t+\tau}
\right)
\right].
\label{eq:subgoal-loss}
\end{equation}
The SmoothL1 term encourages accurate prediction of the target latent values, while the cosine term encourages alignment in feature direction. The encoder $E$ remains frozen throughout training, so both the predicted subgoals and their supervision targets are defined in the same latent space.

\paragraph{Action prior training.}
After training the subgoal generator, we freeze $G$ and train the action prior on the subgoals it predicts. This matches the condition encountered at test time, where the action prior receives predicted rather than ground-truth subgoals, and therefore reduces the misalignment between training and planning. For a recorded action sequence $a_{1:\tau}^{*}$, the action prior is trained with
\begin{equation}
\mathcal{L}_{q}
=
-\log
q\left(
a_{1:\tau}^{*}
\mid
h_t,z_g,\Delta,\tau,z_s^{t,\Delta,\tau}
\right)
+
\lambda_q
\min_m
\operatorname{MAE}
\left(
\mu_m,
a_{1:\tau}^{*}
\right).
\label{eq:option-loss}
\end{equation}
The negative log-likelihood term fits the Gaussian mixture distribution to the expert action sequences $a_{1:\tau}^{*}$, while the second term encourages at least one mixture component mean to remain close to the demonstrated sequence. The Mean Absolute Error (MAE) is computed over the active, normalized action coordinates. Only the action prior is updated in this stage. Additional training and sampling details are provided in Appendices~\ref{app:implementation} and~\ref{app:libero-training}.

\paragraph{Online planning.}
At test time, a query is associated with an initial goal horizon $H$ and an action-duration schedule $(\tau_1,\ldots,\tau_R)$, where $R$ denotes the number of planning stages. At stage $r$, the remaining goal horizon is
\begin{equation}
\Delta_r
=
H-\sum_{j<r}\tau_j.
\label{eq:remaining-horizon}
\end{equation}
The agent first encodes the latest observations to construct the current history $h_t$, and then generates the subgoal
$z_s^{t,\Delta_r,\tau_r}$
conditioned on $(h_t,z_g,\Delta_r,\tau_r)$. The action prior subsequently proposes a population of $\tau_r$-step candidate sequences conditioned on the same information and the generated subgoal. We initialize this population with the mean of the highest-weight GMM component together with $K-1$ samples drawn from the learned mixture distribution, where the candidate population size $K$ is independent of the number of mixture components $M$.

For each candidate $a_{1:\tau_r}$, the frozen world model predicts its terminal latent $F(h_t,a_{1:\tau_r})$, which is scored against the current subgoal $z_s^{t,\Delta_r,\tau_r}$. After action search, the agent executes the resulting $\tau_r$-step sequence open-loop, receives new observations, updates the history, and proceeds to the next planning stage with a newly generated subgoal. The final goal image and its latent representation $z_g$ remain fixed throughout the episode. When $\Delta_r\leq\tau_r$, the final goal already lies within the current planning horizon, so we directly use $z_g$ in place of the intermediate subgoal for both action generation and candidate evaluation. Additional details are provided in Appendices~\ref{app:protocol} and~\ref{app:libero-protocol}.

\section{Experiments}

\newcommand{\PBXXV}{89.3 $\pm$ 1.2}
\newcommand{\PBL}{56.0 $\pm$ 8.0}
\newcommand{\PBLXXV}{20.7 $\pm$ 4.2}
\newcommand{\PBC}{8.7 $\pm$ 3.1}
\newcommand{\PBCXXV}{8.7 $\pm$ 8.1}
\newcommand{\PBCL}{4.7 $\pm$ 4.6}

\newcommand{\PTXXV}{94.0 $\pm$ 3.5}
\newcommand{\PTL}{81.3 $\pm$ 3.1}
\newcommand{\PTLXXV}{81.3 $\pm$ 4.2}
\newcommand{\PTC}{72.7 $\pm$ 9.0}
\newcommand{\PTCXXV}{68.7 $\pm$ 3.1}
\newcommand{\PTCL}{64.7 $\pm$ 9.5}

\newcommand{\PFarXXV}{91.3 $\pm$ 3.1}
\newcommand{\PFarL}{57.3 $\pm$ 8.1}
\newcommand{\PFarLXXV}{18.7 $\pm$ 11.5}
\newcommand{\PFarC}{14.7 $\pm$ 5.0}
\newcommand{\PFarCXXV}{16.0 $\pm$ 4.0}
\newcommand{\PFarCL}{10.7 $\pm$ 6.4}

\newcommand{\PGenXXV}{90.7 $\pm$ 1.2}
\newcommand{\PGenL}{75.3 $\pm$ 9.5}
\newcommand{\PGenLXXV}{70.7 $\pm$ 6.1}
\newcommand{\PGenC}{72.7 $\pm$ 7.6}
\newcommand{\PGenCXXV}{66.7 $\pm$ 7.6}
\newcommand{\PGenCL}{58.7 $\pm$ 7.6}

\newcommand{\PTopXXV}{55.3 $\pm$ 11.5}
\newcommand{\PTopL}{42.0 $\pm$ 6.9}
\newcommand{\PTopLXXV}{35.3 $\pm$ 4.6}
\newcommand{\PTopC}{28.7 $\pm$ 2.3}
\newcommand{\PTopCXXV}{24.0 $\pm$ 10.0}
\newcommand{\PTopCL}{16.0 $\pm$ 5.3}

\newcommand{\PFinalCostXXV}{94.0 $\pm$ 3.5}
\newcommand{\PFinalCostL}{55.3 $\pm$ 9.9}
\newcommand{\PFinalCostLXXV}{23.3 $\pm$ 12.2}
\newcommand{\PFinalCostC}{14.0 $\pm$ 8.0}
\newcommand{\PFinalCostCXXV}{13.3 $\pm$ 3.1}
\newcommand{\PFinalCostCL}{6.7 $\pm$ 7.0}

\newcommand{\CBXXV}{66.7 $\pm$ 4.6}
\newcommand{\CBL}{56.0 $\pm$ 8.7}
\newcommand{\CBLXXV}{66.7 $\pm$ 10.3}
\newcommand{\CBC}{58.7 $\pm$ 9.5}
\newcommand{\CBCXXV}{45.3 $\pm$ 8.1}
\newcommand{\CBCL}{20.7 $\pm$ 5.8}

\newcommand{\CTXXV}{98.7 $\pm$ 1.2}
\newcommand{\CTL}{76.0 $\pm$ 5.3}
\newcommand{\CTLXXV}{86.0 $\pm$ 4.0}
\newcommand{\CTC}{85.3 $\pm$ 1.2}
\newcommand{\CTCXXV}{77.3 $\pm$ 3.1}
\newcommand{\CTCL}{67.3 $\pm$ 3.1}

\newcommand{\CFinalCostXXV}{99.3 $\pm$ 1.2}
\newcommand{\CFinalCostL}{74.0 $\pm$ 2.0}
\newcommand{\CFinalCostLXXV}{82.7 $\pm$ 3.1}
\newcommand{\CFinalCostC}{78.7 $\pm$ 3.1}
\newcommand{\CFinalCostCXXV}{59.3 $\pm$ 11.0}
\newcommand{\CFinalCostCL}{36.7 $\pm$ 3.1}

\newcommand{\CFarXXV}{98.0 $\pm$ 0.0}
\newcommand{\CFarL}{74.0 $\pm$ 3.5}
\newcommand{\CFarLXXV}{83.3 $\pm$ 4.2}
\newcommand{\CFarC}{80.0 $\pm$ 2.0}
\newcommand{\CFarCXXV}{58.7 $\pm$ 10.3}
\newcommand{\CFarCL}{41.3 $\pm$ 2.3}

\newcommand{\CGenXXV}{91.3 $\pm$ 1.2}
\newcommand{\CGenL}{63.3 $\pm$ 5.0}
\newcommand{\CGenLXXV}{76.0 $\pm$ 4.0}
\newcommand{\CGenC}{70.7 $\pm$ 13.3}
\newcommand{\CGenCXXV}{59.3 $\pm$ 11.7}
\newcommand{\CGenCL}{51.3 $\pm$ 4.2}

\newcommand{\CTopXXV}{88.7 $\pm$ 6.4}
\newcommand{\CTopL}{69.3 $\pm$ 3.1}
\newcommand{\CTopLXXV}{60.7 $\pm$ 8.3}
\newcommand{\CTopC}{75.3 $\pm$ 4.2}
\newcommand{\CTopCXXV}{60.7 $\pm$ 4.2}
\newcommand{\CTopCL}{1.3 $\pm$ 2.3}

\newcommand{\APBPL}{56.0 $\pm$ 8.0}
\newcommand{\APBPCL}{4.7 $\pm$ 4.6}
\newcommand{\APGPL}{75.3 $\pm$ 9.5}
\newcommand{\APGPCL}{58.7 $\pm$ 7.6}
\newcommand{\APTPL}{42.0 $\pm$ 6.9}
\newcommand{\APTPCL}{16.0 $\pm$ 5.3}
\newcommand{\APMPL}{81.3 $\pm$ 3.1}
\newcommand{\APMPCL}{64.7 $\pm$ 9.5}

\newcommand{\VSFXXV}{76.0}
\newcommand{\VSFT}{86.0}
\newcommand{\VSLXXV}{64.0}
\newcommand{\VSLM}{72.0}
\newcommand{\VSCXXV}{62.0}
\newcommand{\VSCM}{80.0}
\newcommand{\VSCLXXV}{48.0}
\newcommand{\VSCLM}{60.0}
\newcommand{\VSHFXXV}{80.0}
\newcommand{\VSHFT}{84.0}
\newcommand{\VSOHCF}{84.0}
\newcommand{\VSOHFC}{70.0}
\newcommand{\VSHFCoarseFirst}{78.0}
\newcommand{\VSHFMiddleLong}{76.0}
\newcommand{\VSHHundredAllTwentyFive}{64.0}
\newcommand{\VSHHundredShortTail}{70.0}
\newcommand{\VSHOneTwentyFiveCoarseFirst}{66.0}
\newcommand{\VSHOneTwentyFiveFineFirst}{68.0}
\newcommand{\VSHOneFiftyCoarseFirst}{52.0}
\newcommand{\VSHOneFiftyFineFirst}{56.0}

\begin{figure}[t]
  \centering
  \begin{subfigure}[t]{0.45\textwidth}
    \centering
    \includegraphics[width=\linewidth]{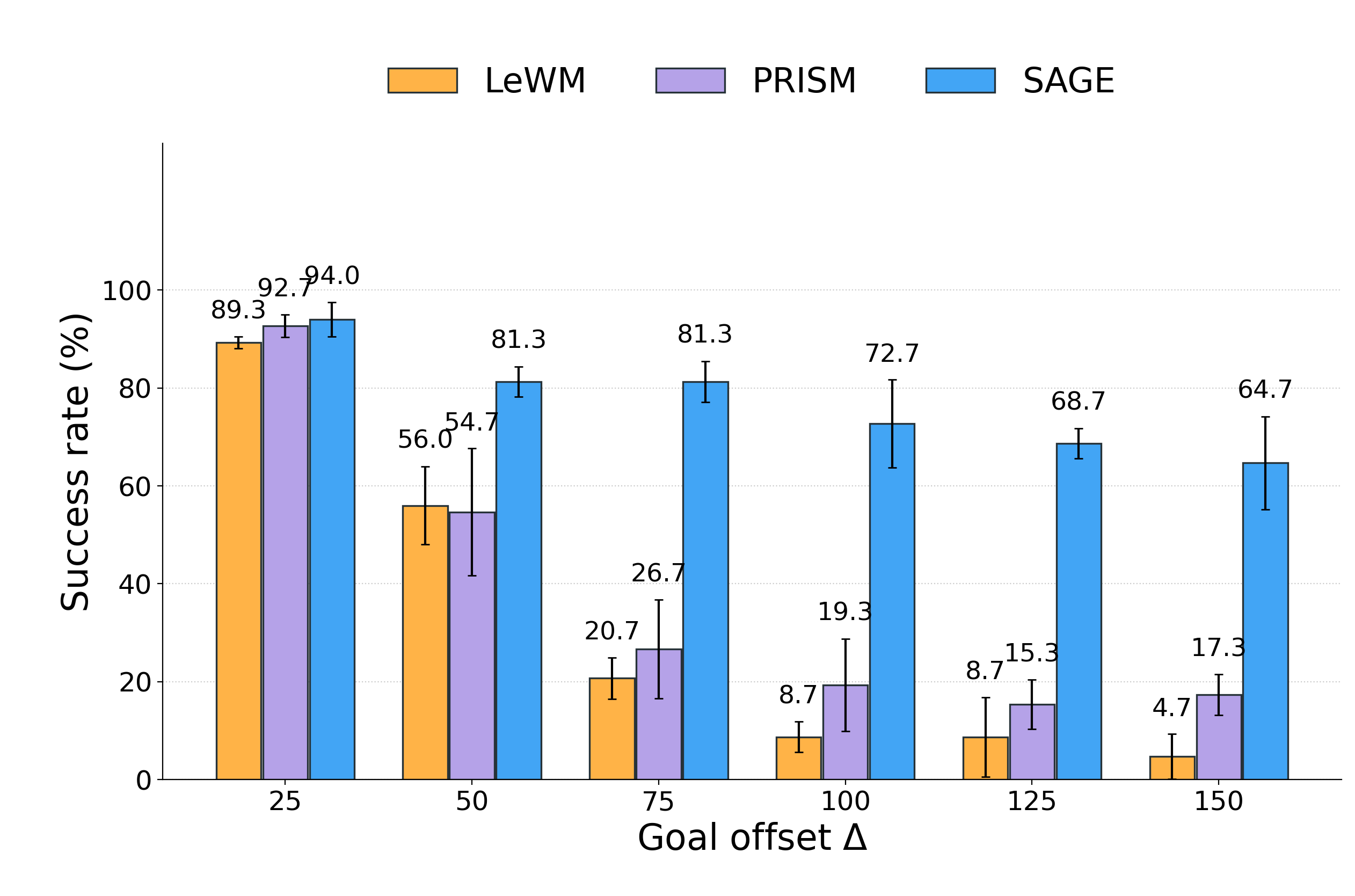}
    \caption{PushT.}
    \label{fig:main-results-pusht}
  \end{subfigure}
  \begin{subfigure}[t]{0.45\textwidth}
    \centering
    \includegraphics[width=\linewidth]{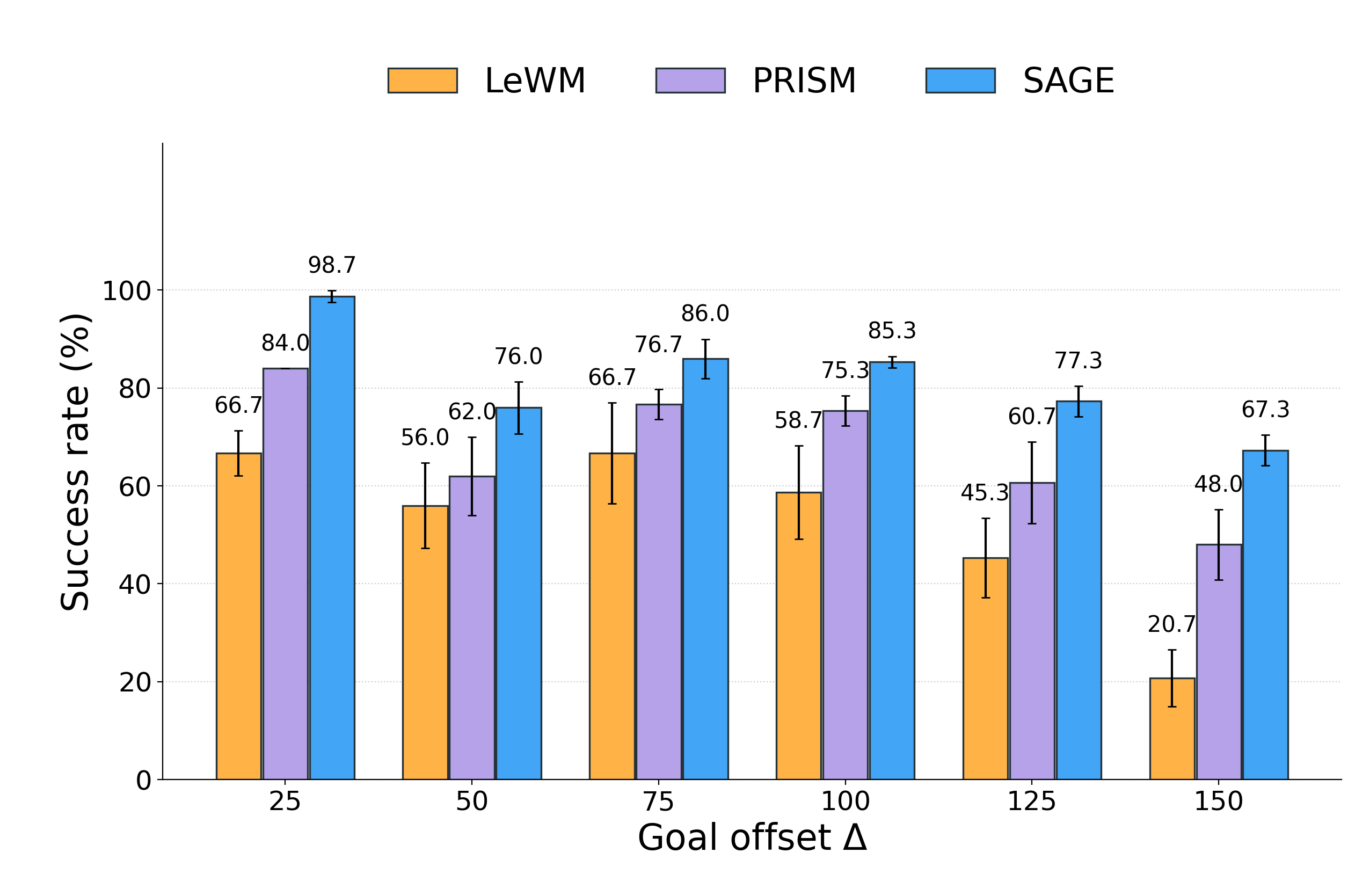}
    \caption{OGBench Cube.}
    \label{fig:main-results-cube}
  \end{subfigure}
  \caption{Success across goal offsets. Bars show the mean over the three fixed
  held-out manifests indexed by seeds $\{32,42,52\}$; error bars show sample
  standard deviation. All methods share the frozen LeWM checkpoint, 300 initial
candidates, and 30 search rounds. Base LeWM and SAGE share the execution
schedule; PRISM uses fixed 25-step chunks.}
  \label{fig:main-results}
\end{figure}

\subsection{Experimental Setup}

\paragraph{Benchmarks.}
We evaluate SAGE under multiple planning tasks. Following prior latent world-model works such as LeWM \citep{lewm} and DINO-WM \citep{dinowm2025}, we first evaluate on PushT and OGBench Cube, which provide controlled goal-reaching benchmarks for studying planning across different target horizons. In both benchmarks, each query pairs the current observation with a future frame from a held-out trajectory, which serves as the image goal, and we evaluate goal offsets \(H\in\{25,50,75,100,125,150\}\). We further evaluate SAGE on LIBERO, a more realistic embodied manipulation benchmark with multi-stage tasks, richer object interactions, and more structured task semantics. This allows us to examine whether the same planning framework can extend from standard latent world-model benchmarks to more challenging full-episode manipulation.

\paragraph{Baselines and evaluation protocol.}
The main experiments in this paper use LeWM \citep{lewm} as the latent world model, with its image encoder and action-conditioned latent dynamics kept frozen throughout SAGE training and evaluation. SAGE only trains the subgoal generator and action prior on top of this fixed latent space. We primarily compare SAGE against Base LeWM, which uses the original Gaussian action proposals, and PRISM \citep{prism2026}, which introduces a learned action proposal prior. We further conduct matched component ablations to isolate the effects of generated subgoals, subgoal-conditioned proposals, the planning target, and world-model search. We report the mean and sample standard deviation over three fixed groups of 50 queries. To test whether SAGE depends on a particular latent representation, we additionally evaluate the same framework with a frozen DINO-WM in Appendix \ref{app:dino-wm}. Additional dataset, execution-schedule, and planning details are provided in the appendix.

\subsection{Long-Horizon Planning Performance}

We first evaluate whether SAGE improves planning as the target moves farther into the future. Figure~\ref{fig:main-results} compares SAGE with Base LeWM and PRISM across different goal horizons on PushT and OGBench Cube. All matched LeWM methods use the same frozen world model, evaluation queries, candidate budget, and search budget, so the comparison isolates the effect of the planning strategy.

As shown in Figure \ref{fig:main-results}, SAGE consistently improves over Base LeWM across the evaluated horizons, with the advantage becoming more pronounced for distant goals. On PushT, Base LeWM achieves 89.3\% success at $H=25$, but drops rapidly to 4.7\% at $H=150$. In contrast, SAGE maintains 94.0\% success at $H=25$ and 64.7\% at $H=150$. A similar trend is observed on OGBench Cube, where SAGE improves success from 20.7\% to 67.3\% at $H=150$. These results show that the performance of the standard planner degrades rapidly as the goal horizon increases, whereas SAGE remains substantially more robust to distant goals.

The short-horizon setting also helps separate the roles of the two components in SAGE. At $H=25$, the final goal already lies within a single planning stage, so SAGE directly uses the final-goal latent without generating an intermediate subgoal. Its improvement in this regime therefore comes primarily from the learned action proposal distribution. As the horizon increases, intermediate subgoals are additionally introduced to guide local planning, leading to increasingly larger gains over Base LeWM. This suggests that learned proposals improve local candidate quality, while subgoal decomposition becomes particularly important for long-horizon planning.

\begin{table*}[t]
\centering
\caption{
Component analysis of SAGE on PushT and OGBench Cube.
We independently vary the action proposal, scoring target, and use of world-model search while keeping the frozen LeWM, evaluation queries, and planning budgets fixed.
Success rate (\%) is reported as mean $\pm$ sample standard deviation over fixed manifests with seeds $\{32,42,52\}$.
}
\label{tab:component-ablation}
\small
\setlength{\tabcolsep}{6pt}
\begin{tabular}{@{}lcccccc@{}}
\toprule
Method
& Proposal
& Scoring Target
& WM Search
& $H=50$
& $H=100$
& $H=150$ \\
\midrule

\multicolumn{7}{@{}l}{\textbf{PushT}} \\

Base LeWM
& Gaussian
& Final goal
& Yes
& \PBL
& \PBC
& \PBCL \\

Final-goal Prior
& Final goal
& Final goal
& Yes
& \PFarL
& \PFarC
& \PFarCL \\

Subgoal Scoring
& Gaussian
& Subgoal
& Yes
& \PGenL
& \textbf{\PGenC}
& \PGenCL \\

Final-goal Scoring
& Subgoal
& Final goal
& Yes
& \PFinalCostL
& \PFinalCostC
& \PFinalCostCL \\

Prior Only
& Subgoal
& --
& No
& \PTopL
& \PTopC
& \PTopCL \\

\textsc{SAGE}
& Subgoal
& Subgoal
& Yes
& \textbf{\PTL}
& \textbf{\PTC}
& \textbf{\PTCL} \\

\addlinespace[3pt]
\multicolumn{7}{@{}l}{\textbf{OGBench Cube}} \\

Base LeWM
& Gaussian
& Final goal
& Yes
& \CBL
& \CBC
& \CBCL \\

Final-goal Prior
& Final goal
& Final goal
& Yes
& \CFarL
& \CFarC
& \CFarCL \\

Subgoal Scoring
& Gaussian
& Subgoal
& Yes
& \CGenL
& \CGenC
& \CGenCL \\

Final-goal Scoring
& Subgoal
& Final goal
& Yes
& \CFinalCostL
& \CFinalCostC
& \CFinalCostCL \\

Prior Only
& Subgoal
& --
& No
& \CTopL
& \CTopC
& \CTopCL \\

\textsc{SAGE}
& Subgoal
& Subgoal
& Yes
& \textbf{\CTL}
& \textbf{\CTC}
& \textbf{\CTCL} \\

\bottomrule
\end{tabular}
\end{table*}

\subsection{Understanding the Gains of SAGE}

We next analyze which components of SAGE are responsible for its long-horizon gains. Table~\ref{tab:component-ablation} decomposes SAGE along three dimensions: how action candidates are proposed, which target is used to score their predicted outcomes, and whether world-model search is used for selection and refinement.

\paragraph{Subgoal-conditioned action proposals.}
We next study whether the subgoal should also shape the proposal distribution. \text{Final-goal Prior} learns an action prior conditioned directly on the final goal, whereas SAGE conditions the prior on the generated subgoal. At long horizons, SAGE consistently performs better, showing that simply replacing Gaussian sampling with a learned prior is not sufficient. This comparison supports coupling learned proposals with temporally matched subgoal guidance.

\paragraph{Local subgoal as the scoring target.}
We first isolate the effect of replacing the distant final goal with a generated local subgoal during candidate evaluation. Comparing \text{Base LeWM} with \text{Subgoal Scoring}, both methods use the same Gaussian proposal distribution and world-model search, but differ in the scoring target. This change alone produces a large improvement at long horizons. On PushT at $H=150$, success increases from 4.7\% to 58.7\%. This indicates that directly evaluating local action candidates against a distant final goal provides a weak planning objective, while a temporally matched subgoal offers a much more informative local target.

\paragraph{Consistent targets for proposal and scoring.}
To separate the roles of proposal generation and candidate evaluation, \text{Final-goal Scoring} uses the same subgoal-conditioned proposal as SAGE but scores predicted outcomes against the final goal. At $H=150$, success drops from 64.7\% to 6.7\% on PushT and from 67.3\% to 36.7\% on Cube. This shows that the generated subgoal should not only guide where candidates are sampled, but should also serve as the target used to evaluate their predicted outcomes.

\paragraph{Role of world-model search.}
Finally, \text{Prior Only} removes world-model search and directly executes the action sequence preferred by the learned prior. Its lower performance than SAGE shows that the action prior and world model play complementary roles. The prior concentrates candidate generation on plausible behaviors, while the frozen world model evaluates and refines these candidates based on their predicted outcomes.

\begin{table}[t]
\centering
\caption{
Temporal-scale analysis on PushT at $H=75$.
The first two rows compare fixed option durations with different replanning frequencies,
while the last two rows reverse the same set of durations within a four-stage nominal schedule.
All rows use the same trained SAGE components, frozen LeWM checkpoint, and per-stage CEM budget.
}
\label{tab:temporal-scale}
\small
\setlength{\tabcolsep}{7pt}
\begin{tabular}{@{}lrr@{}}
\toprule
Option schedule & $R$ & Success (\%) \\
\midrule
$25\times3$ & 3 & \VSFXXV \\
$15\times5$ & 5 & \textbf{\VSFT} \\
\addlinespace
$25+20+15+15$ & 4 & \textbf{\VSOHCF} \\
$15+15+20+25$ & 4 & \VSOHFC \\
\bottomrule
\end{tabular}
\end{table}

\subsection{Planning Across Temporal Scales}

SAGE supports multiple candidate durations through the temporal condition $\tau$, allowing the same trained subgoal generator and action prior to operate at different temporal scales. We use PushT at $H=75$ to study how both the duration and ordering of planning options affect performance.

As shown in Table~\ref{tab:temporal-scale}, using shorter options can improve long-horizon planning when accompanied by more frequent feedback and replanning. Replacing three 25-step options with five 15-step options improves success from \VSFXXV\% to \VSFT\%. This comparison reflects the trade-off between longer open-loop commitment and more frequent opportunities to observe the environment, regenerate a subgoal, and replan.

Beyond the overall option length, the ordering of temporal scales also matters. The schedules $25+20+15+15$ and $15+15+20+25$ contain exactly the same option durations within four nominal planning stages, yet achieve \VSOHCF\% and \VSOHFC\% success, respectively. These results highlight the importance of the temporal execution schedule, including duration ordering and the continuation cadence. Additional schedule comparisons are provided in Appendix \ref{app:temporal-schedules}.

\begin{figure}[htbp]
  \centering
  \includegraphics[width=.9\linewidth]{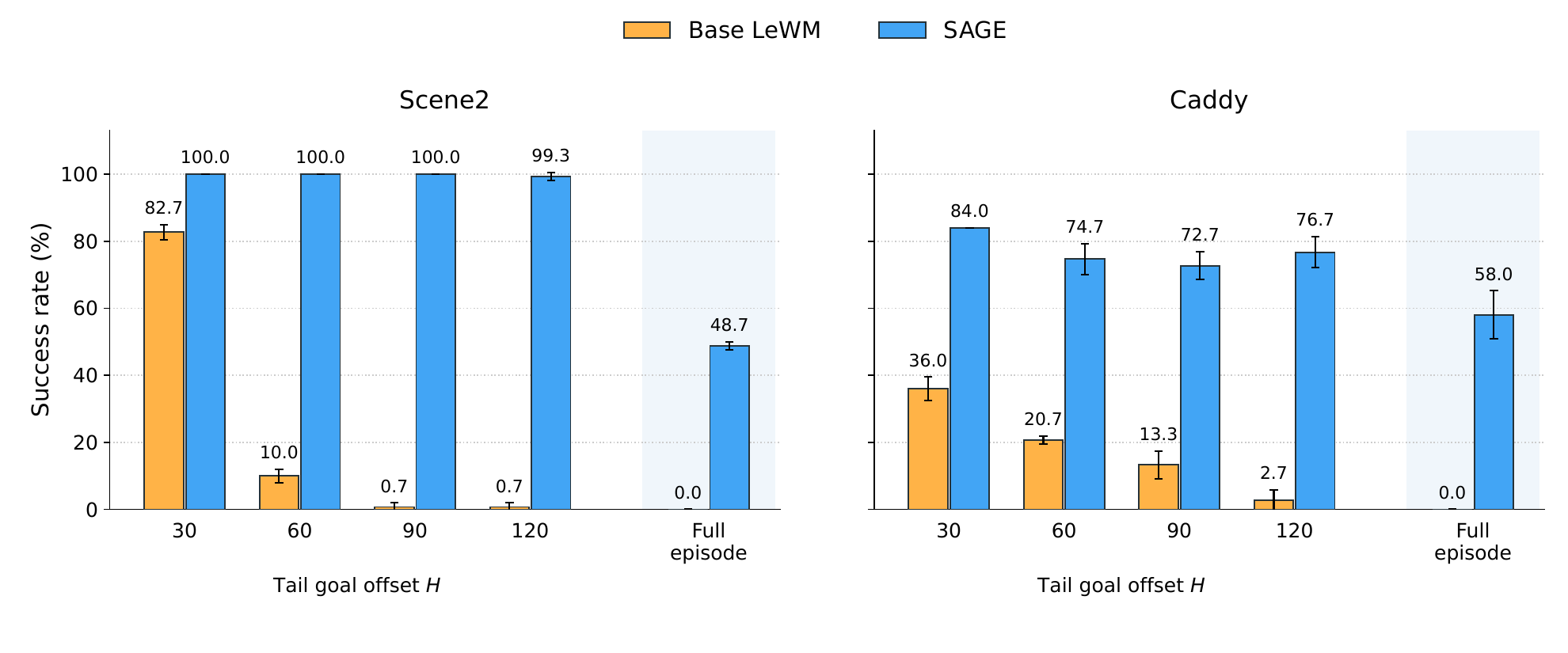}
  \caption{LIBERO official success over 150 queries per condition.
  Bars show mean $\pm$ sample standard deviation over three 50-query groups,
  task-stratified for Caddy (Appendix~\ref{app:libero-protocol}); error bars
  are clipped at zero. Shaded columns show full episodes.
  Both methods share the backbone and execution budgets, including up to
  eight additional 15-step options.}
  \label{fig:libero-main}
\end{figure}

\subsection{Scaling to Long-Horizon Robotic Manipulation}

PushT and OGBench Cube provide controlled settings for studying goal-conditioned planning across different temporal horizons. We further evaluate SAGE on LIBERO to examine whether the same planning framework can extend to more realistic embodied manipulation, where tasks involve multi-stage interactions with different objects and substantially richer task structure.

\paragraph{Data collection and evaluation protocol.}
To support latent world-model training on LIBERO, we construct a reusable closed-loop oracle to collect additional interaction trajectories while preserving the original environments, task definitions, and official success predicates. Each evaluated suite contains 10,000 trajectories, split into 8,000 training, 1,000 validation, and 1,000 test episodes. We consider two evaluation settings. In {tail-goal} planning, the agent starts from a recorded intermediate state $H$ steps before the end of a held-out trajectory and uses the final observation as the image goal. In {full-episode} planning, the agent starts from the beginning of the trajectory and plans toward the same final-image goal over the complete manipulation episode.

\paragraph{From local goal reaching to full-episode planning.}
Figure~\ref{fig:libero-main} shows that SAGE provides substantial gains across tail-goal queries and full-episode evaluation. On Scene2, Base LeWM achieves 82.7\% success at a 30-step tail offset but rapidly degrades for longer-horizon queries and achieves 0\% success from full-episode starts. In contrast, SAGE maintains near-perfect success across the evaluated tail offsets and reaches 48.7\% success on full episodes. A similar trend is observed on Caddy, where SAGE achieves 58.0\% full-episode success compared with 0\% for Base LeWM.

These results show that the benefit of subgoal-conditioned planning extends to more realistic robotic tasks. By repeatedly generating local targets and action proposals throughout the episode, SAGE enables the frozen latent world model to compose multiple local planning decisions over substantially longer and more structured manipulation tasks.

\section{Conclusion}

In this paper, we focus on a key limitation of latent world-model planning: as the goal moves farther into the future, randomly initialized action search becomes increasingly ineffective and the final goal provides weak guidance for local decisions. To address this limitation, we introduce SAGE, which combines a latent subgoal generator that specifies what the agent should achieve locally with a subgoal-conditioned action prior that proposes plausible action sequences toward that target. The frozen world model then evaluates and refines these candidates against the same subgoal, allowing proposal generation and model-based evaluation to operate at a matched temporal scale. Experiments on PushT and OGBench Cube show substantial improvements in long-horizon planning, while component analyses demonstrate that local subgoals, subgoal-conditioned proposals, and world-model search provide complementary benefits. We further extend latent world-model planning to full-episode LIBERO manipulation, where SAGE enables substantially stronger performance on more structured multi-stage tasks. Overall, our results suggest that long-horizon planning depends not only on the quality of the world model, but also on providing it with appropriate local targets and high-quality action candidates to evaluate.

\clearpage
\bibliography{references}
\bibliographystyle{iclr2027_conference}

\clearpage
\appendix
\section{Appendix}
\subsection{Temporal Schedule Analysis}
\label{app:temporal-schedules}

The time-conditioned model supports different planning schedules. We
compare repeated 25-step commitments, repeated 15-step commitments, and
schedules that mix the two. Every stage receives a generated target and
produces an action option at the duration prescribed by that schedule.

\begin{table}[h]
\centering
\caption{Variable-duration planning on PushT, evaluated on a dedicated fixed
held-out manifest indexed by seed 42 ($n{=}50$). Rows share the same generator,
option prior, LeWM checkpoint, and per-stage CEM budget. 
$R$ counts nominal planning stages. After this schedule, the planner repeats its final option duration
until success or the $2H$ execution budget is exhausted.}
\label{tab:variable-schedule}
\small
\begin{tabular}{@{}llrr@{}}
\toprule
Goal horizon & Option schedule & $R$ & Success (\%) \\
\midrule
50 & $25+25$ & 2 & \VSHFXXV \\
50 & $15+15+20$ & 3 & \textbf{\VSHFT} \\
50 & $20+15+15$ & 3 & \VSHFCoarseFirst \\
50 & $15+20+15$ & 3 & \VSHFMiddleLong \\
\addlinespace
75 & $25\times3$ & 3 & \VSFXXV \\
75 & $15\times5$ & 5 & \textbf{\VSFT} \\
75 & $25+20+15+15$ & 4 & \VSOHCF \\
75 & $15+15+20+25$ & 4 & \VSOHFC \\
\addlinespace
100 & $25\times4$ & 4 & \VSHHundredAllTwentyFive \\
100 & $15\times5+25$ & 6 & \VSHHundredShortTail \\
100 & $25+20+20+20+15$ & 5 & \VSLXXV \\
100 & $15+20+20+20+25$ & 5 & \textbf{\VSLM} \\
\addlinespace
125 & $25\times5$ & 5 & \VSCXXV \\
125 & $15\times7+20$ & 8 & \textbf{\VSCM} \\
125 & $25\times2+15\times5$ & 7 & \VSHOneTwentyFiveCoarseFirst \\
125 & $15\times5+25\times2$ & 7 & \VSHOneTwentyFiveFineFirst \\
\addlinespace
150 & $25\times6$ & 6 & \VSCLXXV \\
150 & $15\times10$ & 10 & \textbf{\VSCLM} \\
150 & $25\times3+15\times5$ & 8 & \VSHOneFiftyCoarseFirst \\
150 & $15\times5+25\times3$ & 8 & \VSHOneFiftyFineFirst \\
\bottomrule
\end{tabular}
\end{table}

The schedule diagnostic uses a dedicated query manifest, distinct from the
seed-42 main-table manifest. Reusing the integer seed does not identify the
same queries across these two constructions. Its H100 rows compare two
five-option orderings; the main-table H100 schedule remains
$15\times5+25$ (six options).

More frequent feedback improves success in this comparison. At $H{=}125$,
five 25-step commitments reach $62\%$. Seven 15-step
commitments followed by a 20-step option reach $80\%$. The $R$ column makes the
trade-off explicit: more stages bring more observations, more local-target
queries, and more CEM searches. These rows compare complete temporal policies,
including their feedback cadence and planning compute.

Duration order also matters. At $H{=}75$, $25+20+15+15$ reaches $84\%$,
while $15+15+20+25$ reaches $70\%$. Both schedules contain the same durations
and each use four nominal CEM planning stages before continuation. At $H{=}100$, reversing the order gives $72\%$,
compared with $64\%$ for the first order. Changing duration order changes the
trajectory seen by later planning stages. A future duration selector could
use the current state and remaining horizon to choose the next option length.

\subsection{Generalization Across World Models}
\label{app:dino-wm}

We also train SAGE's generator and prior on the latent space of a frozen
DINO-WM \citep{dinowm2025}. Base DINO-WM uses the same queries and search
budgets. This comparison tests the planner with a second backbone; the
LeWM and DINO-WM checkpoints were trained independently.

\begin{table}[htbp]
\centering
\caption{DINO-WM success (\%). Both benchmarks use 150 queries per horizon,
$K{=}128$ and six CEM rounds. Cube's generator and prior use full
low-dimensional observations; PushT's use visual conditioning.}
\label{tab:dinowm-main}
\small
\setlength{\tabcolsep}{4pt}
\begin{tabular}{@{}llrrrrrr@{}}
\toprule
Benchmark & Method & H25 & H50 & H75 & H100 & H125 & H150 \\
\midrule
PushT & Base DINO-WM & 62.7 & 12.7 & 8.0 & 2.0 & 2.0 & 1.3 \\
& \textsc{SAGE} & \textbf{70.0} & \textbf{32.0} & \textbf{27.3} & \textbf{10.7} & \textbf{8.7} & \textbf{8.7} \\
\addlinespace
Cube & Base DINO-WM & 65.3 & 49.3 & 74.0 & 63.3 & 43.3 & 22.7 \\
& \textsc{SAGE} & \textbf{100.0} & \textbf{91.3} & \textbf{94.0} & \textbf{97.3} & \textbf{94.7} & \textbf{88.0} \\
\bottomrule
\end{tabular}
\end{table}

SAGE improves over Base DINO-WM at every reported horizon on both tasks
(Table~\ref{tab:dinowm-main}), using the same planning procedure as with
LeWM. Appendix~\ref{app:dinowm-cube} gives the component comparison and
backbone-specific settings.

\subsection{PushT and Cube: Environment and Dataset Details}
PushT \citep{diffusionpolicy} is a planar contact-rich manipulation task in which a circular pusher
moves a T-shaped object to a target pose. We use the expert demonstration
dataset and episode split distributed with the LeWM evaluation stack. OGBench
Cube is a robot manipulation task from OGBench \citep{ogbench}; its expert
dataset contains 10,000 episodes of 200 environment steps. Training windows
come exclusively from train episodes, and online evaluation starts from held-out
test states.

\begin{table}[h]
\centering
\caption{Dataset and evaluation summary. Splits are defined at the episode
level before training windows are constructed.}
\label{tab:dataset-summary}
\small
\begin{tabular}{@{}lrrrrl@{}}
\toprule
Task & Train eps. & Val eps. & Test eps. & Action dim. & Success criterion \\
\midrule
PushT & 14,948 & 1,868 & 1,869 & 2 & position $<20$, angle $<\pi/9$ \\
Cube & 8,000 & 1,000 & 1,000 & 5 & target-block distance $<0.04$ m \\
\bottomrule
\end{tabular}
\end{table}

\subsection{PushT and Cube: LeWM Evaluation Protocol}
\label{app:protocol}
An evaluation key consists of
\[(\text{benchmark},\text{split},H,n,\text{sample seed}).\]
For each benchmark, horizon, and sample seed in $\{32,42,52\}$, the sample
builder draws one fixed manifest of $n{=}50$ valid held-out start--goal pairs.
Each record stores an episode id, start frame, goal frame, goal offset, record
id, and a SHA-256 hash of its semantic fields. All methods at a given
benchmark, horizon, and seed consume the same manifest.

PushT evaluates $H\in\{25,50,75,100,125,150\}$ with a $2H$ environment-step
budget; Cube uses an $H$-step budget. Base LeWM and our planner use 300
candidates, 30 CEM rounds, 30 elites, and variance scale 1.0. Main results and
the matched component studies aggregate the same three fixed manifests.

Base LeWM and SAGE use the same nominal execution schedules:
$25$ for $H=25$, $25+25$ for $H=50$, $15\times5$ for $H=75$,
$15\times5+25$ for $H=100$, $15\times7+20$ for $H=125$, and
$15\times10$ for $H=150$. Each horizon has three fixed groups of 50 queries,
indexed by seeds 32, 42, and 52. Reported standard deviations are across
evaluation groups, not independently trained models.

PRISM's original planner uses MPPI. We evaluate its released state-conditioned
proposal head with CEM on PushT and Cube, using 300 candidates and 30 rounds
per decision. It retains fixed 25-step prediction and execution chunks, so
its execution cadence differs from Base and SAGE at longer horizons. In a
separate matched-query PushT H150 comparison, original MPPI achieved 10/100
successes and the CEM variant 17/100.

On PushT, the listed schedule covers the nominal $H$ steps. If the task is
not yet successful, the controller repeats the schedule's final option
duration until the $2H$ budget is exhausted. At elapsed step $s$, its
remaining-duration input is $\max(H-s,\tau)$ for the current option duration
$\tau$, so this input never becomes negative. The far-goal image remains
fixed throughout. This continuation rule applies to the matched Base and
SAGE controllers.

\begin{table}[h]
\centering
\caption{Training and online-planning alignment.}
\label{tab:protocol-matching}
\small
\begin{tabular}{@{}lll@{}}
\toprule
Field & Training tuple & Online planning stage \\
\midrule
History & observations through $t$ & observations through current time \\
Far goal & latent at $t+\Delta$ & query goal with remaining $\Delta_r$ \\
Local target & latent at $t+\tau$ & generator prediction for $\tau_r$ \\
Action target & expert actions over $\tau$ & sampled option over $\tau_r$ \\
Planning target & observed future latent at $t+\tau$ & predicted subgoal or final goal \\
Execution & duration label $\tau$ & execute $\tau_r$ environment steps \\
\bottomrule
\end{tabular}
\end{table}

The planning-target row describes subgoal supervision and its use in search;
the world model remains frozen during SAGE training. When $\Delta_r\leq\tau_r$,
the final-goal latent replaces the predicted subgoal.

\subsection{PushT and Cube: Architecture and Training Details}
\label{app:implementation}
The frozen LeWM encoder produces spatial history and far-goal tokens.
The original PushT components also use current \texttt{state} and
\texttt{proprio} fields (11 coordinates); Cube uses the current
28-dimensional \texttt{observation}, including robot and object state.
Neither uses additional low-dimensional goal-state inputs. Visual-only
variants omit these fields. We project the low-dimensional input separately. Typed
embeddings distinguish history, far goal, state, and time variables. The
subgoal generator uses a four-layer Transformer decoder with width 512 and
eight heads. Goal-shaped queries predict a residual added to the far-goal
latent. It contains 20.71M trainable parameters and optimizes SmoothL1 plus
cosine loss on frozen LeWM targets, with cosine weight 0.1 and SmoothL1
transition parameter 0.05.

The action generator uses a three-layer, width-512, eight-head Transformer
decoder and an eight-mode trajectory GMM. Learned action queries cover the
longest supported option, and the duration mask activates one query per five
environment actions. It contains 13.61M trainable parameters and optimizes
trajectory mixture negative log likelihood plus a best-component normalized
action MAE term weighted by 0.05. Component covariances are diagonal and
learned; predicted log standard deviations are clipped to $[-5,1]$.
Per-query heads predict action means and log standard deviations. Masked
mean pooling of the active query outputs, followed by a linear head and
softmax, produces the trajectory-level mixture weights.

For the reported PushT model, the generator samples far-goal offsets from
\[
\Delta\in\{15,20,25,30,40,45,50,60,65,75,90,100,115,125,140,150\}
\]
and the action generator uses $\tau\in\{15,20,25\}$ whenever $\tau\leq\Delta$.
Both stages use 400k train examples and 40k validation examples drawn from the
full train-window pool. The pair sampler balances the far-goal offsets and the
valid $(\Delta,\tau)$ combinations. The main action generator receives generated
local targets for every training row (Gen100), matching its online input.
Cube uses the same generator--option interface and the same supported option
durations.

\begin{table}[h]
\centering
\caption{Training configuration for the reported PushT model.}
\label{tab:training-hparams}
\small
\begin{tabular}{@{}lll@{}}
\toprule
Field & Subgoal generator & Option prior \\
\midrule
Architecture & Transformer decoder & Transformer decoder + trajectory GMM \\
Depth / width / heads & $4/512/8$ & $3/512/8$ \\
Mixture modes & -- & 8 \\
Trainable parameters & 20.71M & 13.61M \\
Training / validation rows & 400k / 40k & 400k / 40k \\
Batch size & 128 & 128 \\
Optimizer & AdamW & AdamW \\
Learning rate & $10^{-4}$ & $10^{-4}$ \\
Weight decay & $10^{-4}$ & $10^{-4}$ \\
Gradient clipping & 1.0 & 1.0 \\
Precision & BF16 & BF16 \\
Released checkpoint epoch & 6 & 3 \\
Loss & SmoothL1 + $0.1$ cosine & mixture NLL + $0.05$ best-mode MAE \\
\bottomrule
\end{tabular}
\end{table}

\subsection{PushT and Cube: Planner Definitions}
\begin{itemize}
  \item \textbf{Base LeWM}: Gaussian CEM proposals scored by frozen LeWM.
  \item \textbf{LeWM + Generator}: generated local targets paired with Gaussian
  CEM action search.
  \item \textbf{Far-goal Prior + CEM}: a separately trained, capacity-matched
  action prior receives the far-goal latent in place of the generated local
  target. It predicts the same duration-limited expert action options, without
  a subgoal generator. Its proposals initialize CEM, which scores predicted
  futures against the far goal. The option architecture and supported
  durations are unchanged.
  \item \textbf{Generator + Prior Top}: the highest-weight learned option
  conditioned on the same generated local target and final goal, executed
  without LeWM candidate selection or CEM refinement.
  \item \textbf{\textsc{SAGE} (final-goal cost), PushT and Cube}: the generator and
  subgoal-conditioned prior are unchanged. Only the latent target used by
  LeWM to score candidates is replaced with the final-goal latent. The same
  checkpoints, normalization, execution schedule, 300 candidates, 30 CEM rounds,
  and final elite-mean execution are retained; no model is retrained.
  When the remaining horizon is no greater than the option duration, both
  variants use the true final-goal latent without subgoal prediction. Thus
  their H25 objectives are identical. The Cube H25 ablation is an independent
  rerun (149/150 versus the reported SAGE run's 148/150), not a change in objective.
  \item \textbf{\textsc{SAGE}}: generated local targets, duration-matched
  multimodal options, and LeWM-scored elite CEM.
\end{itemize}

In these PushT/Cube LeWM experiments, the initial population contains the
highest-weight GMM component mean and $K-1$ samples from the mixture.
LeWM scores predicted futures against the generated local target, and CEM fits
the next Gaussian to the 30 lowest-cost samples. The remaining 29 rounds use
the same update as Base LeWM.
The executed option is the mean of the final elite set, including when only
one CEM round is used. LIBERO instead returns the minimum-cost candidate
after one ranking pass (Appendix~\ref{app:libero-protocol}).

\FloatBarrier
\subsection{LIBERO Data Construction}
\label{app:libero-data}

\paragraph{Task families and collection.}
We extend LIBERO \citep{libero2023} with additional oracle trajectories.
Scene2 contains five object-to-basket placement tasks, involving alphabet soup,
butter, milk, orange juice, and tomato sauce. Caddy contains three book-placement
tasks targeting the front, left, and right compartments of a caddy in Study
Scene1. We retain the simulator assets and task predicates. A closed-loop
oracle performs approach, grasp, grasp verification, transport, placement,
release, and retreat, with recovery following disturbances. Task-specific
grasp and placement geometry sets the controller parameters. The oracle is
scripted and does not learn a language-conditioned policy from demonstrations.

\paragraph{Perturbations and retained trajectories.}
The collection policy combines oracle actions with temporally correlated
noise and short random-action interventions, followed by oracle recovery.
The Scene2 bank retains collected episodes with explicit success annotations;
Caddy retains successful episodes. Recorded episodes can contain both
perturbations and recovery. Action-mode annotations distinguish ordinary
oracle actions, noise-perturbed oracle actions, and explicit interventions.
The action prior uses modes 0/1 (oracle and noise-perturbed oracle), excluding
explicit random-action bursts from its supervision.

The Caddy collector samples per-episode noise amplitude as
$\sigma=0.10\operatorname{Beta}(2,5)$, with autoregressive coefficient $0.8$
on the first six action coordinates. Intervention durations range from one
to ten raw steps, sampled by clipping
$1+\lfloor10\operatorname{Beta}(1,1.5)\rfloor$ to this interval. The controller
observes the scene again after disturbances and adjusts its actions;
it does not replay a fixed expert suffix.

\begin{table}[htbp]
\centering
\caption{Training datasets for the reported compact LIBERO models. Counts
refer to stored episodes, not overlapping training windows.}
\label{tab:libero-data}
\small
\begin{tabular}{@{}lrrrr@{}}
\toprule
Suite & Tasks & Train & Validation & Test \\
\midrule
Scene2 & 5 & 8,000 & 1,000 & 1,000 \\
Caddy & 3 & 8,000 & 1,000 & 1,000 \\
\bottomrule
\end{tabular}
\end{table}

\begin{table}[htbp]
\centering
\caption{Recorded raw transitions and mean transitions per episode.}
\small
\begin{tabular}{@{}llrr@{}}
\toprule
Suite & Split & Transitions & Mean length \\
\midrule
Scene2 & Train & 1,882,481 & 235.3 \\
& Validation & 235,346 & 235.3 \\
& Test & 235,168 & 235.2 \\
Caddy & Train & 1,956,197 & 244.5 \\
& Validation & 244,848 & 244.8 \\
& Test & 247,007 & 247.0 \\
\bottomrule
\end{tabular}
\end{table}
Scene2's recorded successful episodes are 7,869/8,000 for training and
985/1,000 for each held-out split. These are collection statistics, not
learned-controller evaluation results. Caddy's successful retained bank is
reported separately from the oracle's attempted-rollout success rate.

\paragraph{Data contribution.}
We provide the expanded interaction dataset and closed-loop collection
pipeline for training latent dynamics on LIBERO. Episode splits and recorded
action/image conventions support training and matched image-goal evaluation.

Splits are defined at the episode level before windows are sampled. Caddy
allocates training episodes as 2,667/2,667/2,666 across its three tasks, with
approximately equal validation and test allocations. Larger Scene2 data
collections used in separate scaling studies do not change the 8,000 training
episodes used by the compact model reported here.

\paragraph{Stored information and privileged access.}
Collection records images, raw seven-dimensional actions, episode identity,
and action-mode annotations. Caddy stores agent-view and wrist-view RGB at
$128\times128$. Simulator and object-state information is available to the
oracle and evaluation diagnostics. The learned generator and prior use
visual latents and timing, with zero low-dimensional input channels.

\subsection{LIBERO Evaluation and Runtime Protocol}
\label{app:libero-protocol}

\paragraph{Queries and task balance.}
For a reference episode ending at index $T$, a tail query begins at $T-H$ and
uses image $o_T$ as the goal. Full-episode queries begin at the episode's
initial state and use the same terminal-image convention. Scene2 uses 30
queries per task and Caddy uses 50 per task, totaling 150 per condition.
Table~\ref{tab:libero-main} reports sample standard deviations of group
success percentages. Scene2 uses the manifest's three fixed 50-query groups.
For Caddy, we partition the existing 150 results into three task-stratified
groups of 50. Each group contains 16 queries from one task and 17 from each
of the other two, rotating the 16-query task across groups. Within each task,
episode identities are ordered by SHA-256 with the fixed salt
\texttt{caddy-stratified-20260916}; outcomes are excluded from the assignment.
Base and SAGE share group membership. This reporting partition is applied
after evaluation and leaves all queries and pooled success rates unchanged.
The error bars describe variation between mixed-task query groups, not
independent training seeds or between-task variation.
Methods share query identities within a comparison. A query already satisfying
the success predicate at handoff is not a planning success.

\paragraph{Duration condition and execution budget.}
For full episodes, let $T$ be the recorded number of actions from the
initial state to the terminal goal image. The nominal execution budget is
$15\lfloor T/15\rfloor$, but the generator and prior use the exact remaining
offset, clamped below at 15: $\max(T-s,15)$ after $s$ executed actions.
The goal image is never moved to the rounded index. The maximum budget is
$15\lfloor T/15\rfloor+120$ actions, including eight recovery options.
All options contain 15 actions, even when fewer than 15 reference steps
remain; execution stops early if the task succeeds. Once $T-s\leq15$,
both action generation and scoring use the final-goal latent directly.
The duration input stays at 15 for subsequent recovery options.
For tail queries, $T$ is replaced by the query offset $H$, which is already
a multiple of 15.

\paragraph{History, execution, and success.}
The controller uses three observations at a five-raw-step interval. Missing
initial history is padded with the initial observation, giving three copies
at episode start. Each decision produces a 15-step option, represented by
three five-action tokens; new observations are incorporated at the next
decision. The final goal remains fixed as the duration condition decreases.
When the remaining duration is at most 15, we use the final goal directly as
the local target. All compared controllers receive
eight extra recovery options after the nominal schedule. These are additional
closed-loop actions, not resets or oracle interventions. We check the official
predicate after raw actions and stop at first success. Endpoint latent distance
and geometric tolerances are diagnostic quantities, not the reported success
criterion. The reference duration and recovery allowance distinguish this
protocol from standard fixed-budget, language-conditioned LIBERO evaluation.

\paragraph{Planner definitions.}
The reported LIBERO SAGE configuration evaluates $K=64$ GMM candidates once
per decision: the highest-weight component mean and 63 samples from the mixture.
The implementation flag \texttt{cem\_rounds=1} denotes this
single ranking pass, not an additional Gaussian refit. Generator + Prior Top
uses the same generated local target and terminal condition but executes the
highest-weight component mean without world-model selection. Base LeWM uses
the terminal goal and a nonlearned Gaussian proposal distribution.
For the reported Base, we use a zero-mean-increment residual random walk
with scale 0.05 in native-range units. Its actions are temporally correlated,
unlike the separate IID Gaussian diagnostics, and it does not use a learned
prior's action normalization. The 300-candidate, 30-round settings above apply only
to the original PushT/Cube LeWM experiments.

\paragraph{State restoration and reproducibility.}
Tail queries require matching controller state and rendered observations as
well as joint positions. We reset and replay the reference prefix in a pinned
simulator runtime to obtain the handoff. Cached snapshots can replace this
replay only if their observations and controller behavior match. Replay
metadata includes episode identity, seed, start/goal indices, and image/state
comparisons. A fallback that drops integration state cannot establish exact
restoration. Prefix actions restore the query history and are never used as
future planner candidates. Full-episode evaluation starts without a prefix.

\subsection{LIBERO Model and Training Configuration}
\label{app:libero-training}

Each suite uses its own compact LeWM trained with two Stage-A epochs followed
by two Stage-B epochs. Stage A jointly learns the visual representation and
action-conditioned dynamics with single-step latent prediction and SIGReg;
one model step spans five environment actions. Stage B freezes the encoder
and projector, retains single-step prediction loss, and adds recursive
prediction loss over five model steps (25 actions), weighted by 0.1.
We freeze the encoder for latent caching and the whole
world model during generator and prior training. Stage-B
predictor weights and the latent cache must share the same encoder coordinate
system. Separate training runs are used for Scene2 and Caddy.

\begin{table}[htbp]
\centering
\caption{Checkpoint-recorded configuration for the LIBERO components.
Pair counts are actual sampler outputs, not nominal command-line caps.}
\label{tab:libero-training}
\small
\begin{tabular}{@{}lrr@{}}
\toprule
Field & Scene2 & Caddy \\
\midrule
LeWM scale / stages & approximately 46M / A2+B2 & approximately 46M / A2+B2 \\
Generator width / depth / heads & 768 / 4 / 8 & 768 / 4 / 8 \\
Generator parameters & 46.09M & 46.09M \\
Prior width / depth / heads & 512 / 3 / 8 & 512 / 3 / 8 \\
Prior distribution & trajectory GMM8 & trajectory GMM8 \\
Low-dimensional component inputs & 0 & 0 \\
Local option length & 15 raw steps & 15 raw steps \\
Generator train / validation pairs & 1,531,174 / 190,766 & 1,605,237 / 200,544 \\
Prior train / validation pairs & 1,651,174 / 205,766 & 1,725,237 / 215,544 \\
Generator training / selected epoch & 6 / 6 & 6 / 5 \\
Prior training / selected epoch & 4 / 4 & 4 / 4 \\
Generator / prior batch size & 512 / 256 & 512 / 256 \\
Optimizer / learning rate & AdamW / $10^{-4}$ & AdamW / $10^{-4}$ \\
Weight decay / gradient clip & $10^{-4}$ / 1.0 & $10^{-4}$ / 1.0 \\
\bottomrule
\end{tabular}
\end{table}

\paragraph{Terminal-conditioned supervision.}
The far goal is the episode's terminal frame, while the target subgoal is
15 steps after the sampled current state. Actual duration offsets include
integer values rather than only multiples of 15; the stored maximum offsets
extend beyond 255. The generator predicts a
residual from the current latent, using SmoothL1 loss (transition parameter
0.05) and cosine loss with weight 0.1. The prior conditions on the generated
local target and terminal goal, and uses an eight-component trajectory GMM.
Each sampled option uses one component shared across all its time steps.
Its loss includes mixture NLL and a
mode-L1 auxiliary term weighted by 0.05. The Caddy checkpoint
selects the prior using validation best-of-128 normalized action L1.

\subsection{LIBERO Main Results}
The numerical results underlying Figure~\ref{fig:libero-main} are given in
Table~\ref{tab:libero-main}.

\begin{table}[htbp]
\centering
\caption{LIBERO image-goal planning success (\%), mean $\pm$ sample standard
deviation over three groups of 50 queries. Scene2 uses its fixed evaluation
groups; each Caddy group mixes all three tasks as described in
Appendix~\ref{app:libero-protocol}. Methods share terminal goals, official success
predicates, and recovery budgets. Full episodes start from the initial state.}
\label{tab:libero-main}
\small
\setlength{\tabcolsep}{3pt}
\begin{tabular}{@{}llrrrrr@{}}
\toprule
& & \multicolumn{4}{c}{Tail query} & Full episode \\
Suite & Method & H30 & H60 & H90 & H120 & \\
\midrule
Scene2 & Base LeWM & 82.7 $\pm$ 2.3 & 10.0 $\pm$ 2.0 & 0.7 $\pm$ 1.2 & 0.7 $\pm$ 1.2 & 0.0 $\pm$ 0.0 \\
& \textsc{SAGE} & \textbf{100.0 $\pm$ 0.0} & \textbf{100.0 $\pm$ 0.0} & \textbf{100.0 $\pm$ 0.0} & \textbf{99.3 $\pm$ 1.2} & \textbf{48.7 $\pm$ 1.2} \\
\addlinespace
Caddy & Base LeWM & 36.0 $\pm$ 3.5 & 20.7 $\pm$ 1.2 & 13.3 $\pm$ 4.2 & 2.7 $\pm$ 3.1 & 0.0 $\pm$ 0.0 \\
& \textsc{SAGE} & \textbf{84.0 $\pm$ 0.0} & \textbf{74.7 $\pm$ 4.6} & \textbf{72.7 $\pm$ 4.2} & \textbf{76.7 $\pm$ 4.6} & \textbf{58.0 $\pm$ 7.2} \\
\bottomrule
\end{tabular}
\end{table}

\subsection{LIBERO Component Results}
\label{app:libero-components}

Caddy uses the same full-window prior for tail queries and full episodes.

\begin{table}[htbp]
\centering
\caption{Complete three-controller LIBERO comparison corresponding to the
main figure. Entries are successful queries out of 150; tied best results are
bold. All rows use terminal goals and eight recovery options.}
\label{tab:libero-components}
\small
\setlength{\tabcolsep}{4pt}
\begin{tabular}{@{}llrrrrr@{}}
\toprule
Suite & Controller & H30 & H60 & H90 & H120 & Full \\
\midrule
Scene2 & Base LeWM & 124 & 15 & 1 & 1 & 0 \\
& Generator + Prior Top & \textbf{150} & \textbf{150} & 148 & \textbf{149} & 53 \\
& SAGE & \textbf{150} & \textbf{150} & \textbf{150} & \textbf{149} & \textbf{73} \\
\addlinespace
Caddy & Base LeWM & 54 & 31 & 20 & 4 & 0 \\
& Generator + Prior Top & 97 & 75 & 80 & 76 & 58 \\
& SAGE & \textbf{126} & \textbf{112} & \textbf{109} & \textbf{115} & \textbf{87} \\
\bottomrule
\end{tabular}
\end{table}

On full episodes, SAGE gains 20 successes over Prior Top
on Scene2 and 29 on Caddy,
equivalent to 13.3 and 19.3 percentage points. On Caddy tail queries, SAGE gains
19.3--26.0 percentage points over Prior Top. The comparison with Base measures the joint effect
of structured proposals and world-model selection.

Caddy tail-query and full-episode results use the same prior trained on all
eligible windows, as listed in Table~\ref{tab:libero-training}. The generator
and LeWM are fixed across these comparisons.

\paragraph{Additional execution budget.}
Table~\ref{tab:caddy-recovery} reports success within the nominal horizon and
after two, four, or eight extra 15-step options. Counts are read from the same
rollouts at option boundaries; they are not separate evaluation runs. Both
controllers have the same allowance, with no resets or oracle interventions.
SAGE improves over Prior Top at every budget. The additional options increase
the gap, measuring the benefit of further closed-loop execution rather than
recovery from an injected disturbance.

\begin{table}[htbp]
\centering
\caption{Caddy success versus additional execution budget. Each entry is
Generator + Prior Top / SAGE successes out of 150. An extra option contains
15 raw actions; zero extra options gives a budget of $H$.}
\label{tab:caddy-recovery}
\small
\begin{tabular}{@{}lrrrr@{}}
\toprule
Horizon & 0 extra & 2 extra & 4 extra & 8 extra \\
\midrule
H30 & 75 / \textbf{87} & 87 / \textbf{117} & 92 / \textbf{122} & 97 / \textbf{126} \\
H60 & 50 / \textbf{64} & 67 / \textbf{95} & 71 / \textbf{104} & 75 / \textbf{112} \\
H90 & 57 / \textbf{66} & 70 / \textbf{98} & 74 / \textbf{103} & 80 / \textbf{109} \\
H120 & 56 / \textbf{71} & 70 / \textbf{100} & 73 / \textbf{109} & 76 / \textbf{115} \\
\bottomrule
\end{tabular}
\end{table}

\subsection{LIBERO Decision Latency}
\label{app:libero-latency}

We time one 15-step decision using the same checkpoints as
Table~\ref{tab:libero-components}, including the full-window Caddy prior.
For each task, we use the first locked $H=120$ query and time both controllers
on the same input, with three warm-up calls and 20 alternating repetitions.
Measurements use an RTX 3090, PyTorch 2.5.1, CUDA 12.1, and CUDA
synchronization. Scene2 and Caddy are measured on separate hosts.
Timing includes input preprocessing, image encoding, subgoal generation,
action proposals, and selection; it excludes model loading, prefix replay,
and simulator stepping.

\begin{table}[htbp]
\centering
\caption{LIBERO latency per 15-step decision (ms). Each entry averages the
per-task medians over five Scene2 tasks or three Caddy tasks.}
\label{tab:libero-latency}
\small
\begin{tabular}{@{}lrrr@{}}
\toprule
Suite & Generator + Prior Top & SAGE & Added latency \\
\midrule
Scene2 & 56.6 & 122.3 & 65.7 \\
Caddy & 41.1 & 89.1 & 48.0 \\
\bottomrule
\end{tabular}
\end{table}

SAGE scores 64 candidates once and executes the minimum-cost option;
\texttt{cem\_rounds=1} does not add a Gaussian refit or elite-mean execution.
The added cost is 48--66\,ms per decision, which supplies 15 raw actions.
These repeated-input measurements quantify controller inference cost, not
full-episode wall-clock time or a real-time hardware guarantee.

\FloatBarrier

\subsection{DINO-WM Cube: Protocol and Component Control}
\label{app:dinowm-cube}
The reported Cube model uses DINOv2-base features \citep{dinov22023} and the frozen world-model checkpoint
from epoch 10 of our Cube training run. Compatible generator and prior checkpoints are selected at
epochs 5 and 4, respectively. Their low-dimensional conditioning uses the full
28-dimensional observation, comprising 19 robot-state and nine object-state
coordinates.
All methods use the same three query manifests (seeds 32/42/52, 50 queries
each), $K{=}128$, six CEM rounds for search-based methods, and the original
Cube duration schedule with an $H$-step execution budget and no recovery.
Base uses independently recorded training-action statistics to define its
standardized Gaussian coordinates; zero in this coordinate system does not
mean zero physical action. Its statistics do not come from a prior checkpoint.

\begin{table}[htbp]
\centering
\caption{DINO-WM Cube component comparison, successes out of 150.}
\small
\begin{tabular}{@{}lrrrrrr@{}}
\toprule
Method & H25 & H50 & H75 & H100 & H125 & H150 \\
\midrule
Base DINO-WM & 98 & 74 & 111 & 95 & 65 & 34 \\
Generator + Prior Top & 137 & 136 & 140 & 146 & 143 & 141 \\
SAGE & 150 & 137 & 141 & 146 & 142 & 132 \\
\bottomrule
\end{tabular}
\end{table}

Base measures Gaussian planning, while Prior Top isolates execution without
world-model selection using the same learned components.
SAGE reaches 100\% at H25 and 88\% at H150 on this backbone.
\FloatBarrier
\subsection{DINO-WM PushT Configuration}
The PushT experiment uses the frozen PushT DINO-WM checkpoint with a
six-epoch visual subgoal generator and a four-epoch visual GMM prior. Both
components omit low-dimensional conditioning. All three evaluation groups
use $K=128$, six CEM rounds, 16 elites, action blocks of five raw steps,
four history frames, and a $2H$ environment budget. The Base draws a Gaussian
in the planner's normalized action coordinates; the prior checkpoint supplies
normalization metadata only, not its predicted means or candidates.
The selected generator and prior are shared by all six horizons.

\begin{table}[htbp]
\centering
\caption{DINO-WM PushT: successes out of 150 on the matched manifests.}
\small
\begin{tabular}{@{}lrrrrrr@{}}
\toprule
Method & H25 & H50 & H75 & H100 & H125 & H150 \\
\midrule
Base DINO-WM & 94 & 19 & 12 & 3 & 3 & 2 \\
Generator + Prior Top & 37 & 17 & 6 & 2 & 1 & 1 \\
SAGE & \textbf{105} & \textbf{48} & \textbf{41} & \textbf{16} & \textbf{13} & \textbf{13} \\
\bottomrule
\end{tabular}
\end{table}

\subsection{Complete Visual-Only Results}
\label{app:visual-full}
The low-dimensional input $x$ contains state observations supplied by the
environment, separate from the image latents. Cube's original inputs include
object state as well as robot proprioception. We retrain the generator and
action prior without these inputs, keeping the frozen LeWM and evaluation
protocol fixed. Table~\ref{tab:visual-only} compares the two settings.
Visual-only SAGE retains most of the long-horizon performance on PushT and
remains above Base LeWM on both benchmarks.

\begin{table}[htbp]
\centering
\caption{SAGE with and without low-dimensional conditioning. Success (\%),
pooled over three manifests (150 queries per cell). State-assisted denotes
the original benchmark-specific inputs, not proprioception only.}
\label{tab:visual-only}
\small
\begin{tabular}{@{}llrr@{}}
\toprule
Benchmark & Conditioning & H25 & H150 \\
\midrule
PushT & State-assisted & \textbf{94.0} & \textbf{64.7} \\
& Visual-only & 93.3 & 61.3 \\
Cube & State-assisted & \textbf{98.7} & \textbf{67.3} \\
& Visual-only & 96.7 & 56.0 \\
\bottomrule
\end{tabular}
\end{table}

These experiments use the original LeWM backbones. Both the subgoal
generator and action prior have zero low-dimensional input channels,
including goal-state channels. Generator width/depth/heads are 512/4/8;
the GMM8 prior uses 512/3/8. The generator is the six-epoch checkpoint
for each benchmark; the PushT and Cube priors are from epochs three and
four, respectively. The schedule, candidates, CEM rounds, and query
identities follow the original LeWM study.

\begin{table}[htbp]
\centering
\caption{Visual-only SAGE: each group contains 50 queries. Pooled rows
give percentages over all 150 queries, not training-seed averages.}
\small
\begin{tabular}{@{}llrrrrrr@{}}
\toprule
Benchmark & Query group & H25 & H50 & H75 & H100 & H125 & H150 \\
\midrule
PushT & 32 & 48 & 46 & 41 & 38 & 37 & 30 \\
& 42 & 45 & 39 & 37 & 39 & 33 & 31 \\
& 52 & 47 & 40 & 39 & 31 & 32 & 31 \\
& Pooled (\%) & 93.3 & 83.3 & 78.0 & 72.0 & 68.0 & 61.3 \\
\addlinespace
Cube & 32 & 47 & 36 & 44 & 43 & 40 & 28 \\
& 42 & 49 & 37 & 40 & 40 & 37 & 25 \\
& 52 & 49 & 42 & 42 & 37 & 37 & 31 \\
& Pooled (\%) & 96.7 & 76.7 & 84.0 & 80.0 & 76.0 & 56.0 \\
\bottomrule
\end{tabular}
\end{table}

\subsection{Additional RoboTwin Manipulation}
\label{app:robotwin-a2b}
We evaluate RoboTwin 2.0 \citep{robotwin2025} \texttt{place\_a2b\_left} with a stride-one
observation dataset and its matched approximately 89M LeWM trained with
A2+B2. The subgoal generator and GMM action prior use the same encoder
coordinates as this world model. Queries are anchored to the recorded
terminal frame, with three disjoint groups of 50 episodes per horizon.
Each controller receives the same terminal image, 15-step commitment,
and eight recovery options (120 additional raw actions). SAGE ranks
64 candidates in one pass; Prior Top executes the highest-weight GMM
component mean. Base uses the current observed robot state to anchor a
Gaussian residual random walk with scale 0.05 of the action range, without
expert action statistics or a future reference-action center. Success is
the environment task predicate, checked during execution.

\begin{table}[htbp]
\centering
\caption{RoboTwin A2B terminal-tail evaluation, successes out of 150.}
\small
\begin{tabular}{@{}lrrrr@{}}
\toprule
Controller & H30 & H60 & H90 & H120 \\
\midrule
Base LeWM & 0 & 0 & 11 & \textbf{19} \\
Generator + Prior Top & 13 & 15 & 6 & 0 \\
SAGE & \textbf{23} & \textbf{21} & \textbf{21} & 18 \\
\bottomrule
\end{tabular}
\end{table}
This experiment uses the same subgoal--action interface with a different
robot and action convention. It measures terminal-tail execution, not
full-episode completion.

\FloatBarrier
\subsection{Training-Seed Replication}
\label{app:training-seeds}
We repeat PushT component training with two additional optimization seeds
(248 and 348), holding the frozen LeWM, dataset split, and training-pair
sampling fixed. Each run uses 400,000 training pairs for each component,
the epoch-6 generator, and the epoch-3 GMM prior, following the original
component recipe. We evaluate each trained pair on the same three evaluation
seeds (32, 42, and 52), with 50 queries per seed and horizon, using
$K=300$ and 30 CEM iterations. Only the component optimization seed changes
across runs; the world-model checkpoint and training split stay fixed.

\begin{table}[htbp]
\centering
\caption{PushT training-seed replication. Entries are success percentages
(successful queries out of 150). Evaluation seeds are shared across rows.}
\label{tab:training-seeds}
\small
\begin{tabular}{@{}lrrr@{}}
\toprule
Component training run & H50 & H100 & H150 \\
\midrule
Original & 81.3 (122/150) & 72.7 (109/150) & 64.7 (97/150) \\
Seed 248 & 82.0 (123/150) & 68.7 (103/150) & 63.3 (95/150) \\
Seed 348 & 80.0 (120/150) & 72.0 (108/150) & 59.3 (89/150) \\
\bottomrule
\end{tabular}
\end{table}
Across the three trained component pairs, mean success is 81.1\% at H50,
71.1\% at H100, and 62.4\% at H150.
\FloatBarrier

\subsection{Gaussian Baseline Conventions}
\label{app:gaussian-definitions}
The action coordinate system determines what a Gaussian proposal samples. Zero mean in
normalized coordinates, zero native action, and zero increments around a
current-state anchor are different distributions. Table~\ref{tab:gaussian}
records the conventions used by the reported experiments. Baselines do not
use future expert actions as proposal centers. Past reference actions, when
used to restore a query, serve only to initialize the environment.
Search-based methods share the query and execution budget within
each reported comparison; the prior-only control has no search budget.

\begin{table}[htbp]
\centering
\caption{Benchmark-specific Gaussian Base and execution conventions.
``One pass'' denotes candidate ranking without iterative CEM refitting.}
\label{tab:gaussian}
\footnotesize
\setlength{\tabcolsep}{4pt}
\begin{tabular}{@{}p{0.19\linewidth}p{0.39\linewidth}p{0.12\linewidth}p{0.22\linewidth}@{}}
\toprule
Setting & Gaussian convention & Candidates / rounds & Execution budget \\
\midrule
LeWM PushT / Cube & Initial $\mathcal N(0,I)$ in planner action coordinates; independent samples, zero candidate included, no warm start & 300 / 30 & SAGE schedule; PushT $2H$; Cube $H$ \\
\addlinespace
DINO-WM PushT & Zero-mean Gaussian in normalized action coordinates; checkpoint statistics only & 128 / 6 & $2H$, original schedule \\
\addlinespace
DINO-WM Cube & Zero-mean Gaussian in independently computed training-action standardized coordinates & 128 / 6 & $H$, original schedule \\
\addlinespace
LIBERO Scene2 / Caddy & Zero-mean-increment residual random walk; fixed scale 0.05 in native-range units & 64 / one pass & 15-step options; eight recovery options \\
\addlinespace
RoboTwin A2B & Current-observed-state anchor plus cumulative Gaussian residuals; fixed range fraction 0.05 & 64 / one pass & 15-step options; eight recovery options \\
\bottomrule
\end{tabular}
\end{table}
For standardized coordinates $u$, a zero-mean Gaussian maps to native
actions as $a=\mu_{\rm train}+\sigma_{\rm train}\odot u$; it is not a
zero-native-action distribution. For a residual random walk, independent
increments are accumulated along the option before action-range constraints
are applied. The corresponding temporal correlation is part of the baseline
definition and does not depend on a learned action prior.
\FloatBarrier

\subsection{Estimated Goal Horizon}
\label{app:estimated-horizon}
We test whether SAGE can use an estimated goal offset in place of the
reference offset. A visual estimator takes three frozen LeWM latents,
spaced five steps apart, and the goal latent. We concatenate these latents
and the goal-minus-current difference, then use a two-hidden-layer MLP
(width 512, 0.76M parameters) to predict a distribution over the generator's
16 training offsets, from 15 to 150 steps. Inputs are standardized using
training statistics. The estimator uses no low-dimensional state inputs.

Training uses 400,000 pairs and validation uses 40,000 pairs, balanced over
offsets and separated by episode. We minimize cross-entropy with AdamW
(learning rate $3\times10^{-4}$, weight decay $10^{-4}$, batch size 1,024)
for up to 20 epochs, selecting the lowest validation cross-entropy.
Validation classification accuracy is 70.8\%; the posterior-mean offset
has a mean absolute error of 5.11 steps, with 90.7\% of estimates within
15 steps of the reference offset. These labels describe recorded trajectory
timing, not the minimum time needed to reach the goal.

For online evaluation, we use the epoch-6 generator and epoch-3 prior from
component training seed 248 (Appendix~\ref{app:training-seeds}), without
retraining either component. At each planning call, the estimator predicts
the offset from the current visual history and goal. Its rounded posterior
mean replaces the reference offset in both learned components and in the
decision to use the final goal directly. We rerun the reference-offset
control with the same checkpoints and queries. Both conditions retain
300 candidates, 30 CEM rounds, the prescribed execution schedule, and the
$2H$ environment budget. The schedule and budget still depend on the
evaluation horizon; this experiment tests replacing the reference offset
in policy conditioning, rather than removing all horizon information.

\begin{table}[htbp]
\centering
\caption{PushT with reference and estimated goal offsets. Success (\%),
mean $\pm$ sample standard deviation over the same three 50-query manifests
(seeds 32, 42, and 52). Both rows use the same trained SAGE components.}
\label{tab:estimated-horizon}
\small
\begin{tabular}{@{}lrrr@{}}
\toprule
Goal offset & H50 & H100 & H150 \\
\midrule
Reference & $80.7\pm8.3$ & $67.3\pm11.7$ & $63.3\pm4.6$ \\
Estimated & $78.0\pm10.6$ & $70.0\pm5.3$ & $68.0\pm10.6$ \\
\bottomrule
\end{tabular}
\end{table}
We observe no consistent loss of success when replacing the reference
offset. The H150 increase is concentrated in one query group, so these
results support the use of estimated offsets without establishing an
advantage over reference offsets.
\FloatBarrier

\FloatBarrier

\clearpage
\subsection{Qualitative Held-Out Rollouts}
\begin{figure*}[!t]
  \centering
  \includegraphics[width=0.98\textwidth]{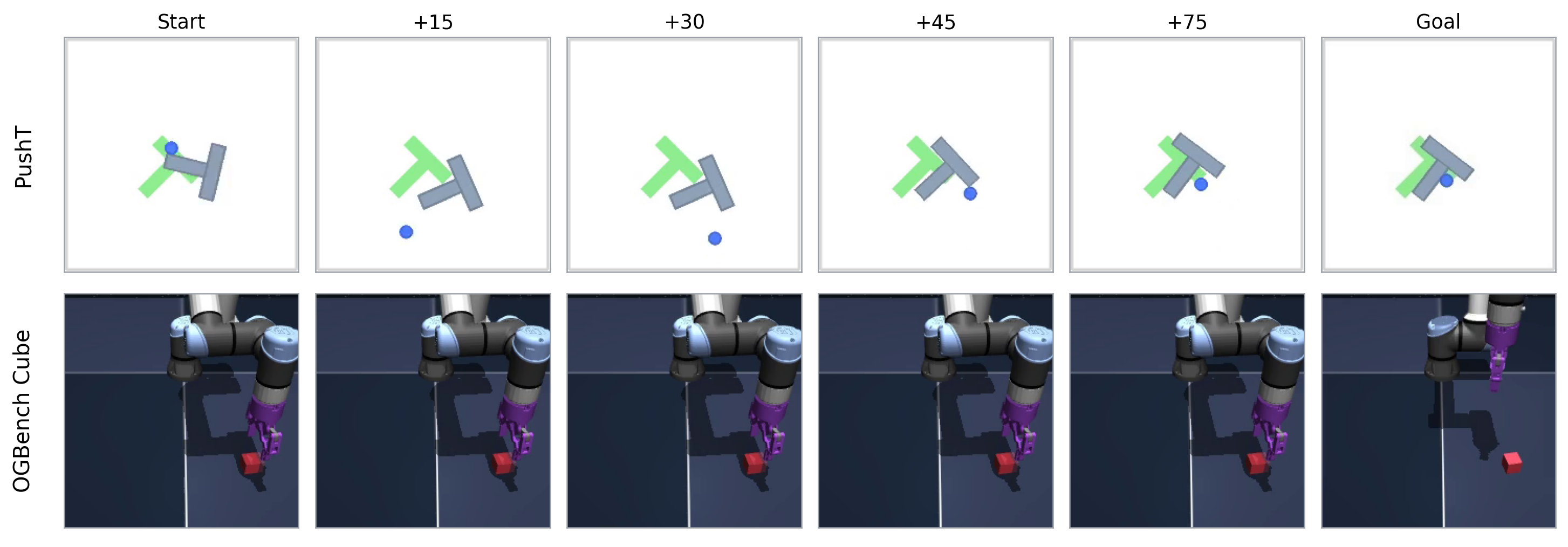}
  \caption{Representative successful held-out rollouts at $H{=}75$.
  Each row shows the online observation at the start, after 15, 30, 45, and
  75 environment steps, followed by the query goal. The intermediate frames
  come from the planner's own closed-loop rollout; they do not come from the
  offline demonstration.}
  \label{fig:rollout-examples}
\end{figure*}

\subsection{Per-Seed Main Results}
\begin{table}[h]
\centering
\caption{PushT success rate (\%) on the fixed held-out manifests indexed by
seeds $\{32,42,52\}$.}
\label{tab:pusht-per-seed}
\small
\begin{tabular}{@{}llrrrrrr@{}}
\toprule
Seed & Method & H25 & H50 & H75 & H100 & H125 & H150 \\
\midrule
32 & Base LeWM & 90 & 64 & 22 & 12 & 4 & 2 \\
32 & \textsc{SAGE} & 96 & 82 & 80 & 82 & 66 & 72 \\
42 & Base LeWM & 88 & 48 & 16 & 6 & 18 & 2 \\
42 & \textsc{SAGE} & 90 & 78 & 86 & 72 & 68 & 54 \\
52 & Base LeWM & 90 & 56 & 24 & 8 & 4 & 10 \\
52 & \textsc{SAGE} & 96 & 84 & 78 & 64 & 72 & 68 \\
\bottomrule
\end{tabular}
\end{table}

\subsection{Capacity Study}
\label{app:capacity}
We vary the capacity of both learned components while keeping the
temporal interface, GMM mode count, candidate budget, CEM updates, and the
held-out seed-42 manifest fixed. The temporal schedule is unchanged.
The Small generator and prior both use width
256, two decoder layers, and four attention heads; the prior retains eight
mixture components. The Small prior checkpoint is epoch 4, whereas the Main
prior checkpoint is epoch 3. Training budgets therefore differ between the
two configurations. The Small stack retains short-horizon control, while
the Main stack scores higher on the longer planning queries.

\begin{table}[h]
\centering
\caption{PushT capacity study on the fixed seed-42 manifests. Both rows use
the default duration schedule and 300-candidate elite CEM.}
\label{tab:push-capacity}
\small
\resizebox{\linewidth}{!}{%
\begin{tabular}{@{}lrrr rrrrrr@{}}
\toprule
Scale & Generator & Planner & Total & H25 & H50 & H75 & H100 & H125 & H150 \\
& (M) & (M) & (M) & & & & & & \\
\midrule
Small & 3.00 & 2.41 & 5.41 & 92 & 58 & 66 & 70 & 48 & 46 \\
Main & 20.71 & 13.61 & 34.31 & 90 & 78 & 86 & 72 & 68 & 54 \\
\bottomrule
\end{tabular}
}
\end{table}

\end{document}